\documentclass[lettersize,journal]{IEEEtran}
\usepackage{algorithmic}
\usepackage{algorithm}
\usepackage{array}
\usepackage[caption=false,font=normalsize,labelfont=sf,textfont=sf]{subfig}
\usepackage{textcomp}
\usepackage{stfloats}
\usepackage{url}
\usepackage{verbatim}
\usepackage{graphicx}
\usepackage{cite}
\usepackage[caption=false,font=normalsize,labelfont=sf,textfont=sf]{subfig}
\usepackage{amsmath,amssymb,amsfonts}
\usepackage{amsthm}
\usepackage{algorithmic,algorithm}
\usepackage{xcolor,soul}
\usepackage{arcs}
\usepackage{cases}
\newtheorem{theorem}{Theorem}

\newtheorem{lemma}{Lemma}
\newtheorem{proposition}{Proposition}

\newtheorem{assumption}{Assumption}
\newtheorem{rem}{Remark}
\newtheorem{fact}{Fact}
\usepackage{comment}
\usepackage[colorlinks=true,linkcolor=blue, citecolor=green, urlcolor=magenta]{hyperref}

\allowdisplaybreaks 

\begin{document}
\title{Hybrid Feedback Control for Global Navigation with Locally Optimal Obstacle Avoidance in $n$-Dimensional Spaces}
\author{Ishak Cheniouni,~\IEEEmembership{Member,~IEEE}, Soulaimane Berkane,~\IEEEmembership{Senior Member,~IEEE}, Abdelhamid Tayebi,~\IEEEmembership{Fellow,~IEEE}
\thanks{The authors are with the Department of Electrical Engineering, Lakehead University, Thunder Bay, ON P7B 5E1, Canada (e-mail: {\tt\small cheniounii, atayebi@lakeheadu.ca}). S. Berkane is also with the Department of Computer Science and Engineering, University of Quebec in Outaouais, 101 St-Jean Bosco, Gatineau, QC, J8X 3X7, Canada (e-mail: {\tt\small soulaimane.berkane@uqo.ca}).}
}%


\maketitle

\begin{abstract}
We present a hybrid feedback control framework for autonomous robot navigation in $n$-dimensional Euclidean spaces cluttered with spherical obstacles. The proposed approach ensures safe and global navigation towards a target location by dynamically switching between two operational modes: motion-to-destination and locally optimal obstacle-avoidance. It produces continuous velocity inputs, ensures collision-free trajectories and generates locally optimal obstacle avoidance maneuvers. Unlike existing methods, the proposed framework is compatible with range sensors, enabling navigation in both a priori known and unknown environments. Extensive simulations in 2D and 3D settings, complemented by experimental validation on a TurtleBot 4 platform, confirm the efficacy and robustness of the approach. Our results demonstrate shorter paths and smoother trajectories compared to state-of-the-art methods, while maintaining computational efficiency and real-world feasibility.
\end{abstract}

\section{Introduction}
\label{sec:introduction}
\subsection{Motivation}
The challenge of autonomous navigation in obstacle-filled environments remains a critical area in robotics, with broad applications ranging from mobile robots to aerial and underwater vehicles. A key requirement is to ensure safety while reaching the destination from any location in the free space, avoiding randomly distributed obstacles. Existing methods often struggle with trade-offs between safety, optimality, and computational efficiency, especially in unknown or dynamic environments. The seminal artificial potential field approach \cite{khatib} provided an intuitive solution by considering destinations as attractive forces and obstacles as repulsive ones. However, the susceptibility to local minima limits its effectiveness. Navigation functions \cite{k_R_90}, \cite{R_k_92} later addressed this limitation by achieving almost global asymptotic stability (AGAS) in structured environments like sphere worlds. Despite these advancements, incorporating path-length optimality and ensuring global guarantees remain open challenges. Our work is motivated by bridging these gaps to provide a robust framework for navigating high-dimensional spaces safely and efficiently.
\subsection{Related Work}
Autonomous navigation has been extensively studied, with a range of methods addressing the challenges of obstacle avoidance. The artificial potential field (APF) approach \cite{khatib} is among the earliest and most influential methods, offering a simple yet effective framework by modeling the destination as an attractive force and obstacles as repulsive forces. Despite its intuitive appeal, APF suffers from the presence of local minima, which can trap robots and prevent them from reaching their target.
 This issue was addressed by the introduction of the concept of navigation functions (NFs) \cite{k_R_90}, which achieves almost global asymptotic stability (AGAS) of the target location in sphere worlds. The NF-based approach has since been extended to more complex environments as detailed in \cite{R_k_92}. 
In \cite{LoizouNT4, LoizouNT3}, a new NF has been developed in point worlds, enabling autonomous navigation in sphere and star worlds, leading to AGAS guarantees. In \cite{SnsNF6}, a NF-based approach has been proposed for autonomous robot navigation in environments with convex obstacles that are relatively flat with respect to their distance to the target. This limitation was overcome in \cite{KUMAR2022110643}, where AGAS of the target location is achieved in environments with arbitrarily flat ellipsoidal obstacles. Under the same flatness condition as in \cite{SnsNF6}, the authors in \cite{Arslan2019} proposed an almost global reactive sensor-based approach. Their idea consists in enclosing the robot inside a local safe region obtained by the intersection of the hyperplanes separating the robot from the adjacent obstacles. The projection of the target on this convex safe region is considered as an intermediary local destination for the robot. This work has been extended in \cite{vasilopoulos1,Vasilopoulos2} to environments with star-shaped and polygonal obstacles with possible overlap. A control framework unifying control Lyapunov functions and control barrier functions through quadratic programs (QPs) was proposed in \cite{Barrier6,Barrier7}. Despite its efficiency in combining safety and stabilization requirements, this QP-based framework suffers from the presence of stable undesired equilibria \cite{Undesired_CBF_21}. A modified version of the QP-based control was proposed in \cite{Undesired_CBF_21,undesired_CBF_Dimos} to eliminate certain types of undesired equilibria while ensuring local asymptotic stability of the target location. 

\begin{figure}[h!]
    \centering
    \subfloat[]{
        \includegraphics[width=\linewidth]{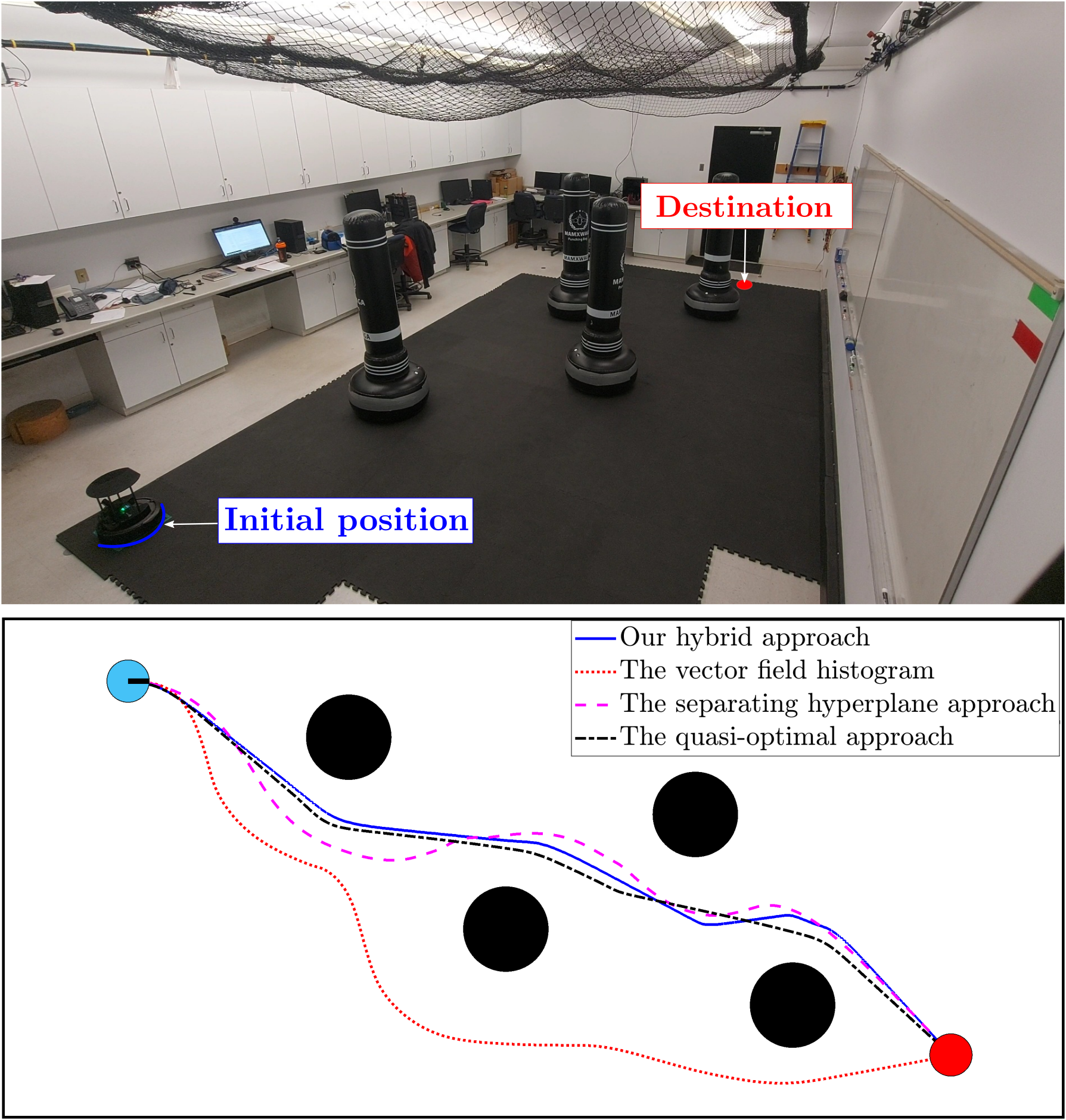}\label{fig:compexp_results}
    }
    \vfill
    \subfloat[]{
        \resizebox{1\linewidth}{!}{
        \begin{tabular}{|c||c||c||c||c|}
            \hline
            \textbf{Algorithm} & \textbf{SH} & \textbf{VFH} & \textbf{QO} & \textbf{Hybrid approach} \\
            \hline
            \textbf{Execution time} & 13.21$ms$ & 1.69$ms$ & 0.43$ms$ & 0.88$ms$ \\
            \hline
            \textbf{Path length} & 7.8405$m$ & 8.1877$m$ & 7.4541$m$ & 7.5584$m$ \\
            \hline
        \end{tabular}
        }
        \label{table_exp_time_comp}
    }
    \caption{ 
    (a) Example navigation scenario in a priori unknown environment, showing the robot's trajectory generated by the proposed hybrid feedback control approach (blue) compared to alternative methods. (b) Performance comparison highlighting the path length and computational efficiency of the proposed approach.  The proposed approach generates paths similar to our previously proposed quasi-optimal (QO) approach \cite{Ishak2023} while avoiding the issue of undesired equilibria in QO approach. The details of this experiment are reported in Section \ref{experimental_validation}. The complete experiment can be visualized in the video available online \url{https://youtu.be/KzUNLwQ5lMo}. 
    }
    \label{fig:combined_results}
\end{figure}

While these approaches have made significant strides in addressing safety and stability in navigation, achieving both path-length optimality and global asymptotic stability (GAS) in arbitrary, high-dimensional environments with complex obstacle configurations remains a key unresolved challenge. Path-length optimality, in particular, is one of the main objectives of path planning algorithms, whereas feedback-based navigation approaches, such as those mentioned above, generally do not prioritize the shortest path. Moreover, many of these methods, such as those based on navigation functions or quadratic programming (QP), rely on constructing at least a local representation of the environment, which can be computationally intensive and impractical for real-time applications in dynamic or unknown settings. 
Reactive path planners, on the other hand, offer an alternative by relying solely on local sensing to plan paths in real-time. Bug algorithms exemplify this category of navigation strategies, operating under two primary modes of motion: motion toward the target and boundary following. The first two bug algorithms, Bug1 and Bug2, introduced in \cite{Bug1}, relied on contact sensors for obstacle detection. Later variants, such as VisBug \cite{Bug2} and TangentBug \cite{Kamon1998TangentBugAR}, incorporated range sensors for improved performance. However, the efficiency of bug algorithms strongly depends on the geometry of the workspace, as highlighted in \cite{ng2007performance}. Despite their computational simplicity and adaptability, these reactive planners often generate suboptimal trajectories and are typically constrained to 2D settings, limiting their applicability in more complex environments. The velocity obstacle (VO) approach is another reactive method, first introduced for dynamic obstacles with known velocities \cite{VO_1} and later extended to multi-agent navigation \cite{VO_survey}. VO and its variants guarantee collision-free motion within a finite time horizon but do not provide convergence guarantees to the target.

To address path-length optimality, path planning algorithms typically rely on constructing grids or graphs based on exact or approximate representations of the configuration space \cite{lavalle_2006}. These algorithms then compute the shortest path using search techniques such as Dijkstra's algorithm \cite{Dijkstra1959ANO}, A* (A star) algorithm \cite{Astar}, or Theta* (Theta star) algorithm \cite{theta_star}. Combinatorial methods rely on the exact configuration space, leading to complete algorithms and exact shortest paths. One of the earliest combinatorial methods is the visibility graph approach \cite{Nilsson1969AMA}, designed for two-dimensional environments with polygonal obstacles. This approach was optimized in \cite{ROHNERT198671}, called the tangent visibility graph, and then extended in \cite{LAUMOND198741,arimoto} to solve the shortest path problem in two-dimensional environments with curved obstacles. Sampling-based methods do not require an explicit representation of the configuration space but use a sampling scheme to explore the configuration space, resulting in weaker notions of completeness, such as probabilistic completeness. Among this category, one can find single query planning algorithms such as the rapidly exploring random trees (RRTs) \cite{LaValle1998RapidlyexploringRT} or multiple query algorithms such as the probabilistic roadmaps (PRMs) \cite{PRMs}. Although endowed with probabilistic completeness, the RRT and PRM algorithms generally provide non-optimal solutions, while the variants RRT* and PRM*, proposed in \cite{karaman}, guarantee asymptotically optimal solutions. These approaches, however, are computationally expensive and rely heavily on prior knowledge of the environment, limiting their real-time applicability in dynamic scenarios.


In an attempt to incorporate path length optimality in feedback-based approaches, we proposed, in \cite{ACC23,Ishak2023}, a continuous feedback control strategy in sphere worlds generating quasi-optimal trajectories. Unfortunately, sets of non-zero Lebesgue measure, from where the undesired equilibria may be reached, may exist in two-dimensional environments unless some restrictions on the obstacles configuration are enforced. A sensor-based version of this approach, with AGAS guarantees of the target location, has also been proposed in \cite{Ishak2023} to deal with convex obstacles satisfying a curvature condition similar to the one in \cite{SnsNF6,Arslan2019}. A global result is out of reach for the above-cited works, involving feedback-based continuous controllers, due to the topological obstructions pointed out in \cite{k_R_90}. As an alternative, hybrid feedback controllers have been proposed in the literature to achieve global asymptotic stability (GAS) of the target location. The work in \cite{HybBerkaneECC2019} achieves GAS of the target location in Euclidean spaces with a single spherical obstacle. An extension was proposed in \cite{SoulaimaneHybTr} for multiple ellipsoidal obstacles in Euclidean spaces. Recently, hybrid feedback control for safe and global navigation in two-dimensional environments with arbitrary convex obstacles was proposed in \cite{Mayur2022}. A similar control scheme was proposed more recently in \cite{Mayur_non_convex} for environments with non-convex obstacles. Although the above-mentioned hybrid feedback-based approaches provide GAS results, the generated trajectories are not optimal in terms of length. Path-length optimality, along with GAS of the target location, has been achieved in our recent work \cite{cheniouni2024hybrid}, in $n$-dimensional Euclidean spaces but only for a single spherical obstacle. However, achieving both path-length optimality and global asymptotic stability (GAS) for arbitrary numbers of obstacles in general 
$n$-dimensional environments, without restrictive assumptions on obstacle configurations, remains an open challenge, highlighting a critical gap in existing methodologies. 

\subsection{Contributions and Organization of the Paper}
This paper proposes a hybrid feedback control strategy for safe autonomous navigation in $n$-dimensional Euclidean spaces with spherical obstacles. The strategy operates in two distinct modes: the {\it motion-to-destination} mode, where the robot moves directly towards the target when it has a clear line of sight, and the locally-optimal {\it obstacle-avoidance} mode, when the robot does not have a clear line of sight to the target location. The main contributions of the proposed approach are summarized as follows:

\begin{itemize}
    \item \textbf{Safety and global navigation}: The proposed control strategy ensures safe navigation in $n$-dimensional spaces with spherical obstacles while converging to the target location from all initial conditions in the workspace.
    \item \textbf{Continuous control input}: Unlike many hybrid strategies, the proposed hybrid feedback controller produces continuous velocity inputs, enabling smooth robot motion and ensuring practical feasibility for real-world applications.
    \item \textbf{Local optimal avoidance maneuvers}: By dynamically generating shortest-path maneuvers around obstacles, the controller achieves locally optimal navigation.
    \item \textbf{Navigation in unknown environments}: The proposed obstacle-avoidance mechanism is fully implementable using range sensors alone, allowing navigation in both 2D and 3D environments without requiring prior global knowledge. Experimental validation on a TurtleBot 4 platform demonstrates its effectiveness in \textit{a priori} unknown settings, as illustrated in Fig. \ref{fig:combined_results}. \item \textbf{Scalability and computational efficiency}: The control strategy is computationally lightweight and scalable to higher-dimensional spaces, making it well-suited for autonomous systems with limited onboard resources.
\end{itemize}

Our earlier conference paper \cite{cheniouni2024hybrid} is a brief preliminary version of the present work dealing only with a single \textit{a priori} known obstacle. The present work provides a significant extension of the research by addressing the autonomous navigation problem in \textit{a priori} unknown $n$-dimensional environments with multiple obstacles. Unlike \cite{cheniouni2024hybrid}, the present paper provides rigorously proven sensor-based autonomous navigation solutions ensuring both global navigation and locally optimal obstacle-avoidance maneuvers. In addition, we carry out a comprehensive experimental study and a comparison with state-of-the-art approaches, providing stronger validation of the practical applicability of the proposed approach. Links to videos from our experiments and GitHub repository containing the implemented codes, are provided for the readers' reference.

The remainder of this paper is organized as follows:
Section II introduces the preliminaries and notations used throughout the paper. Section III formulates the autonomous navigation problem. Section IV details the proposed control strategy and its key properties, including safety and stability guarantees. Section V discusses the sensor-based implementation of the control strategy for 2D and 3D workspaces. Simulation results are presented in Section VI, and Section VII reports the experimental validation of the approach using a TurtleBot 4 platform. Finally, Section VIII concludes the paper by summarizing the contributions and outlining directions for future work. 
\section{Notations and Preliminaries}
Throughout the paper, $\mathbb{N}$ and $\mathbb{R}$ denote the set of natural numbers and real numbers, respectively. The Euclidean space and the unit $n$-sphere are denoted by $\mathbb{R}^n$ and $\mathbb{S}^n$, respectively. The Euclidean norm of $x\in\mathbb{R}^n$ is defined as $\|x\|:=\sqrt{x^\top x}$ and the angle between two non-zero vectors $x,y\in\mathbb{R}^n$ is given by $\angle (x,y):=\cos^{-1}(x^\top y/\|x\|\|y\|)$. The interior, the boundary, and the closure of a set $\mathcal{A}\subset\mathbb{R}^n$ are denoted by $\mathring{\mathcal{A}}$, $\partial\mathcal{A}$, and $\overline{\mathcal{A}}$, respectively. The relative complement of a set $\mathcal{B}\subset\mathbb{R}^n$ with respect to a set $\mathcal{A}\subseteq\mathbb{R}^n$ is denoted by $\mathcal{A}\setminus\mathcal{B}$. The distance of a point $x\in\mathbb{R}^n$ to a closed set $\mathcal{A}$ is defined as $d(x,\mathcal{A}):=\min\limits_{q\in\mathcal{A}}\|q-x\|$. 
The elementary reflector is defined as $\pi_r(v):=I_n-2vv^\top$ where $I_n\in\mathbb{R}^{n\times n}$ is the identity matrix and $v\in\mathbb{S}^{n-1}$ \cite{Matrix_Analysis_and_Applied_Linear_Algebra}. Therefore, for any vector $x$, the vector $\pi_r(v)x$ corresponds to the reflection of $x$ about the hyperplane orthogonal to $v$. The line passing by two points $x,y\in\mathbb{R}^n$ is defined as $\mathcal{L}(x, y):=\left\{q\in\mathbb{R}^n|q=x+\lambda(y-x),\;\lambda\in\mathbb{R}\right\}$ and the line segment between the points $x,y\in\mathbb{R}^n$ is defined as $\mathcal{L}_s(x, y):=\left\{q\in\mathbb{R}^n|q=x+\lambda(y-x),\;\lambda\in[0,1]\right\}$. We define the ball centered at $x\in\mathbb{R}^n$ and of radius $r>0$ by the set $\mathcal{B}(x,r):=\left\{q\in\mathbb{R}^n|\;\|q-x\| \leq r\right\}$.
Let us define the set $\mathcal{P}_{\Delta}(x,v)=\left\{q\in\mathbb{R}^n|v^\top(q-x)~\Delta~0\right\}$, with $\Delta \in\{=,>,\geq,<,\leq\}$.
The hyperplane passing through $x\in\mathbb{R}^n$ and orthogonal to $v\in\mathbb{R}^n\setminus\{0\}$ is denoted by $\mathcal{P}_{=}(x,v)$. The closed negative half-space (resp. open negative half-space) is denoted by $\mathcal{P}_{\leq}(x,v)$ \big(resp. $\mathcal{P}_{<}(x,v)$\big) and the closed positive half-space (resp. open positive half-space) is denoted by $\mathcal{P}_{\geq}(x,v)$ \big(resp. $\mathcal{P}_{>}(x,v)$\big). A conic subset of $\mathcal{A}\subseteq\mathbb{R}^n$, with vertex $x\in\mathbb{R}^n$, axis $v\in\mathbb{R}^n$, and aperture $2\varphi$ is defined as follows \cite{HybBerkaneECC2019}:
\begin{align}
    \mathcal{C}^{\Delta}_{\mathcal{A}}(x,v,\varphi):=\left\{q\in\mathcal{A}|\|v\|\|q-x\|\cos(\varphi)\Delta v^\top(q-x)\right\},
\end{align}
where $\varphi\in(0,\frac{\pi}{2}]$ and $\Delta\in\left\{\leq,<,=,>,\geq\right\}$,  with $``="$, representing the surface of the cone, $``\leq"$ (resp. $``<"$) representing the interior of the cone including its boundary (resp. excluding its boundary), and $``\geq"$ (resp. $``>"$) representing the exterior of the cone including its boundary (resp. excluding its boundary). The set of vectors parallel to the cone $\mathcal{C}^=_{\mathbb{R}^n}(x,v,\varphi)$ is defined as follows:
    \begin{align}\label{parallel_cone}
        \mathcal{V}(v,\varphi):=\left\{w\in\mathbb{R}^n|\;\;w^\top v=\|w\|\|v\|\cos(\varphi)\right\}.
    \end{align}
We state a property of cones sharing the same vertex as follows:
\begin{lemma}[\cite{HybBerkaneECC2019}]\label{lem1}
Let $c,v_{-1},v_{1}\in\mathbb{R}^n$ such that $\angle(v_{-1},v_{1})=\psi$ where $\psi\in(0,\pi]$. Let $\varphi_{-1},\varphi_{1}\in[0,\pi]$ such that $\varphi_{-1}+\varphi_{1}<\psi<\pi-(\varphi_{-1}+\varphi_{1})$. Then 
\begin{align}
    \mathcal{C}_{\mathbb{R}^n}^{\leq}(c,v_{-1},\varphi_{-1})\cap\mathcal{C}_{\mathbb{R}^n}^{\leq}(c,v_{1},\varphi_{1})=\{c\}.
\end{align}
\end{lemma}%
Finally, a hybrid dynamical system is represented by
\begin{align} {\begin{cases}\dot{X}\in \mathrm{F}(X), &X\in \mathcal {F}\\ X^+\in \mathrm{J}(X), & X\in \mathcal {J} \end{cases}} \label{hyb} \end{align}
 where $X \in \mathbb{R}^n$ is the state, the (set-valued) flow map $\mathrm{F} : \mathbb{R}^n \rightrightarrows \mathbb{R}^n$
and the (set-valued) jump map $\mathrm{J} : \mathbb{R}^n \rightrightarrows \mathbb{R}^n$ govern continuous and discrete evolution,
which can occur, respectively, in the flow set $\mathcal{F} \subset \mathbb{R}^n$ and the jump set
$\mathcal{J} \subset \mathbb{R}^n$. The notions of solution $\phi$ to a hybrid system, its hybrid time domain $\text{dom}\,\phi$, maximal and complete solution can be found in
in \cite[Def. 2.6, Def. 2.3, Def. 2.7, p. 30]{Sanfelice} .
\section{Problem Formulation}
Autonomous navigation in cluttered environments remains a fundamental challenge in robotics, particularly when requiring safety, efficiency, and smooth motion in high-dimensional spaces.
We consider the position $x\in\mathbb{R}^n$ of a point mass vehicle evolving in the Euclidean space of dimension $n\geq2$ punctured by $b\in\mathbb{N}\setminus\{0\}$ balls $\mathcal{O}_k:=\mathcal{B}(c_k,r_k)$ of radius $r_k>0$ and center $c_k\in\mathbb{R}^n$ where $k\in\mathbb{I}:=\{1,\dots,b\}$. The obstacle-free space is, therefore, given by the closed set
\begin{align}\label{freespace}
    \mathcal{X}:=\mathbb{R}^n\setminus\bigcup\limits_{k\in\mathbb{I}}\mathring{\mathcal{O}}_k.
\end{align}
In practical applications, obstacle disjointness is a standard requirement to avoid overlapping regions, which could lead to navigation ambiguities or infeasible maneuvers. Therefore, we impose the following assumption to preserve the spherical nature of obstacles: 
\begin{assumption}\label{as1}
    The obstacles are pairwise disjoint, that is,
 \begin{align}
    \|c_k-c_j\|>r_k+r_j,\;\;\forall k,j\in\mathbb{I},\,i\neq j.
\end{align}
\end{assumption}
We consider a velocity-controlled vehicle such that \( \dot{x} = u \), where \( u \) represents the control policy designed to generate trajectories in constrained $n-$dimensional spaces. This model assumes full actuation, which is a common abstraction for theoretical exploration, though practical implementations may incorporate actuation constraints. The primary objective is to design a feedback control policy for \( u \) that ensures safe and efficient navigation while addressing the following challenges:  
\begin{itemize}
    \item \textbf{Global Goal Reaching}: Steer the vehicle from any initial position \( x_0 \in \mathcal{X} \) to a predefined destination \( x_d \in \mathring{\mathcal{X}} \), ensuring the robot consistently reaches the goal regardless of the initial conditions.  
    \item \textbf{Obstacle Avoidance}: Prevent collisions with all obstacles \( \mathcal{O}_k \), leveraging the obstacle-free space \( \mathcal{X} \) to navigate safely.  
    \item \textbf{Locally Optimal Maneuvers}: Achieve locally shortest feasible paths around obstacles, reflecting efficiency in navigation and minimizing unnecessary detours.  
    \item \textbf{Smoothness}: At each time $t$, the control input \( u(t) \) is continuous, leading to continuously differentiable trajectories suitable for practical deployment.
\end{itemize}
The proposed problem addresses a critical gap in existing navigation frameworks. Many approaches either rely on constructing explicit or local representations of the environment, which may not be feasible in real-time, or fail to guarantee smooth, optimal, and globally converging trajectories in \( n \)-dimensional spaces. By focusing on disjoint spherical obstacles, we establish a mathematically tractable yet practically relevant scenario that enables rigorous analysis of the control strategy. The emphasis on continuous control inputs ensures compatibility with robotic systems where abrupt changes can destabilize dynamics or degrade performance.
This problem formulation sets the stage for developing a novel hybrid feedback strategy that overcomes these limitations, providing a robust solution for autonomous navigation in high-dimensional obstacle-filled environments.

\section{Main Results}\label{section:main_results}
In this section, we present the design of our hybrid controller, demonstrating the forward invariance of the obstacle-free space and the stability of the target location under the proposed control scheme. Furthermore, we establish the continuity of the control input and substantiate the optimality of the obstacle-avoidance maneuvers.
\subsection{Control design}\label{section:control-design}
The proposed hybrid control strategy consists of two operation modes: the \textit{motion-to-destination} mode and the \textit{obstacle-avoidance} mode. We make use of a mode selector $m\in\mathbb{M}:=\{-1,0,1\}$ which refers to the \textit{motion-to-destination} mode if $m=0$, and to the \textit{obstacle-avoidance} mode if $m=\pm 1$. Each mode of operation is activated and deactivated in specific regions--referred to as flow sets and jump sets--based on a carefully designed switching strategy. The proposed hybrid feedback control depends on the current position $x\in\mathbb{R}^n$, the considered obstacle $k\in\mathbb{I}$, and the operation mode $m$, and is defined as follows:
\begin{subequations}\label{hyb_ctrl}
    \begin{align}
    &u(x,k,m)=m^2\alpha(x,k)\mu(x,k,m)\kappa(x,k,m)\nonumber\\
    &\qquad\qquad\qquad\qquad\qquad+(1-m^2\alpha(x,k))u_d(x),\label{hyb_ctrl_L}
    \end{align}
    \begin{align}
    &\qquad\begin{cases}
        \Dot{k}=0,\\
        \Dot{m}=0,
    \end{cases}&&(x,k,m)\in\mathcal{F},\label{jump_dyn_c}\\
    &\qquad\begin{cases}
        k^{+}\in K(x,k,m),\\
        m^{+} \in M(x,k,m)
    \end{cases}&&(x,k,m)\in\mathcal{J},\label{jump_dyn}
    \end{align}
\end{subequations}
where $u_d(x):=-\gamma(x-x_d)$ is the nominal control for the {\it motion-to-destination} mode, $\gamma>0$, $\kappa(x,k,m)$ is the control for the {\it obstacle-avoidance} mode that will be defined in Subsection \ref{projection_mode_design}. The scalar functions $\alpha$ and $\mu$, which will be defined in Subsection \ref{Switching_scheme}, ensure smooth transitions between the operation modes. The dynamics of the obstacle and mode selectors are described in \eqref{jump_dyn_c}-\eqref{jump_dyn} where $K(\cdot)$ is the jump map of the obstacle selector and $M(\cdot)$ is the jump map of the mode selector that will be designed in Subsection \ref{Switching_scheme}. The sets 
\begin{align}\label{sets}
\mathcal{F}:=\bigcup\limits_{m\in\mathbb{M}}\left(\mathcal{F}_m\times\{m\}\right),\;\mathcal{J}:=\bigcup\limits_{m\in\mathbb{M}}\left(\mathcal{J}_m\times\{m\}\right)
\end{align}
are, respectively, the flow and jump sets of the hybrid system where $\mathcal{F}_m$ and $\mathcal{J}_m$ are, respectively, the flow and jump sets of the operation mode $m\in\mathbb{M}$ that will be constructed in Subsections \ref{M-to-D}, \ref{projection_mode_design}. In the following, we first define the necessary sets for our control design. Then, we define the control in each mode and its associated flow and jump sets, and after that, we define the jump maps that govern the transitions between these modes.
\subsubsection{Sets definition}\label{section:sets}
\begin{figure}[h!]
\centering
\includegraphics[width=0.8\columnwidth]{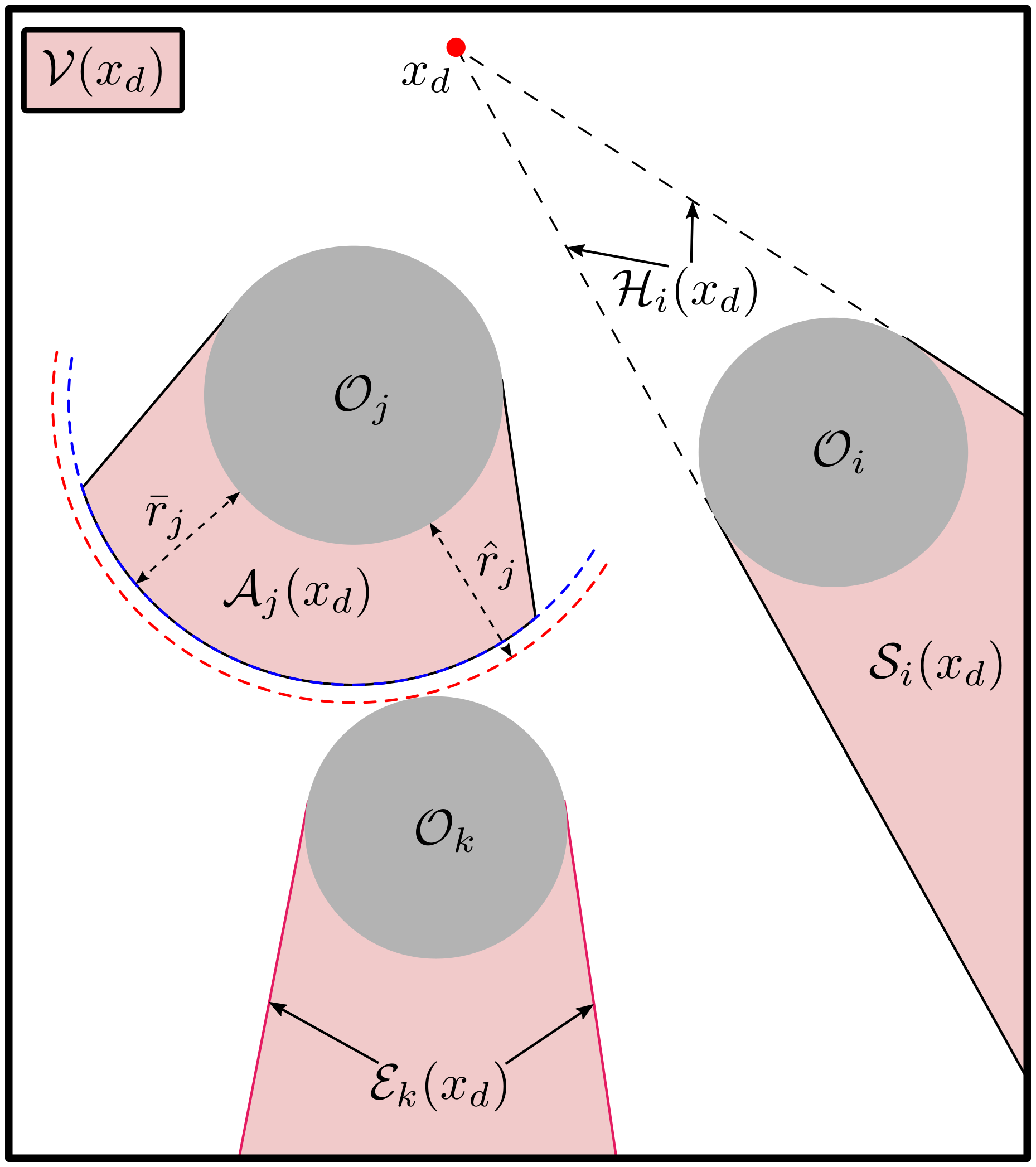}
\caption{2D representation of the sets in Section \ref{section:sets}.}
\label{fig1}
\end{figure}

In this subsection, we define the subsets of the free space that will be used in the design of our control. These subsets are illustrated in Fig. \ref{fig1} and presented as follows:
\begin{itemize}
    \item The \textit{shadow region} of obstacle $\mathcal{O}_k$ is the area hidden by obstacle $\mathcal{O}_k$ from which the vehicle does not have a clear line of sight to the target. It is defined as follows:
    \begin{multline}\label{shadow_region}
        \mathcal{S}_k(x_d):=\bigl\{q\in\mathcal{C}^{\leq}_{\mathcal{X}}(x_d,c_k-x_d,\theta(x_d,k))|\\(c_k-q)^\top(x_d-q)\geq0\bigr\},
    \end{multline}
    where the function $\theta(q,k)\colon\mathcal{X}\to (0,\frac{\pi}{2}]$, $q\mapsto\theta(q,k):=\arcsin(r_k/\|q-c_k\|)$ assigns to each position $q$ of the free space, the half aperture of the cone enclosing obstacle $\mathcal{O}_k$.
    \item The \textit{active region} of obstacle $\mathcal{O}_k$ is defined as follows:
    \begin{align}\label{active_region}
        \mathcal{A}_k(x_d):=\mathcal{S}_k(x_d)\cap\mathcal{B}(c_k,r_k+\Bar{r}_k),
    \end{align}
    where $\Bar{r}_k\in(0,\hat{r}_k)$, $\hat{r}_k=\min\limits_{j\in\mathcal{I}_k(x_d)}\left(\|c_k-c_j\|-r_k-r_j\right)$, and $\mathcal{I}_k(x_d):=\left\{j\in\mathbb{I}|\mathcal{S}_k(x_d)\cap\partial\mathcal{O}_j\neq\varnothing\right\}$ is the set of obstacles hidden (fully or partially) from the destination $x_d$ by obstacle $\mathcal{O}_k$. Note that when $\mathcal{I}_k(x_d)=\varnothing$, $\hat{r}_k=\infty$ and $\mathcal{A}_k(x_d)=\mathcal{S}_k(x_d)$.
    \item The \textit{exit set} of obstacle $\mathcal{O}_k$ is the lateral surface of the {\it active region} and is defined as follows:
    \begin{multline}\label{12}
        \mathcal{E}_k(x_d):=\mathcal{C}^{=}_{\mathcal{X}}(x_d,c_k-x_d,\theta(x_d,k))\cap\mathcal{A}_k(x_d).
    \end{multline}
    \item The \textit{hat} of obstacle $\mathcal{O}_k$ is the upper part of the surface of the cone of vertex $x_d$ enclosing obstacle $\mathcal{O}_k$ and is defined as follows:
    \begin{align}\label{hat}
        \mathcal{H}_k(x_d):=\mathcal{C}^{=}_{\mathcal{X}}(x_d,c_k-x_d,\theta(x_d,k))\setminus\mathcal{E}_k(x_d).
    \end{align}
    \item The \textit{active free space} is defined as
    \begin{align}\label{active_freespace}
        \mathcal{V}(x_d):=\bigcup\limits_{k\in\mathbb{I}}\mathcal{A}_k(x_d).
    \end{align}
\end{itemize}
\subsubsection{Motion-to-destination mode ($m=0$)}\label{M-to-D}
\begin{figure}[h!]
\centering
\includegraphics[width=0.7\columnwidth]{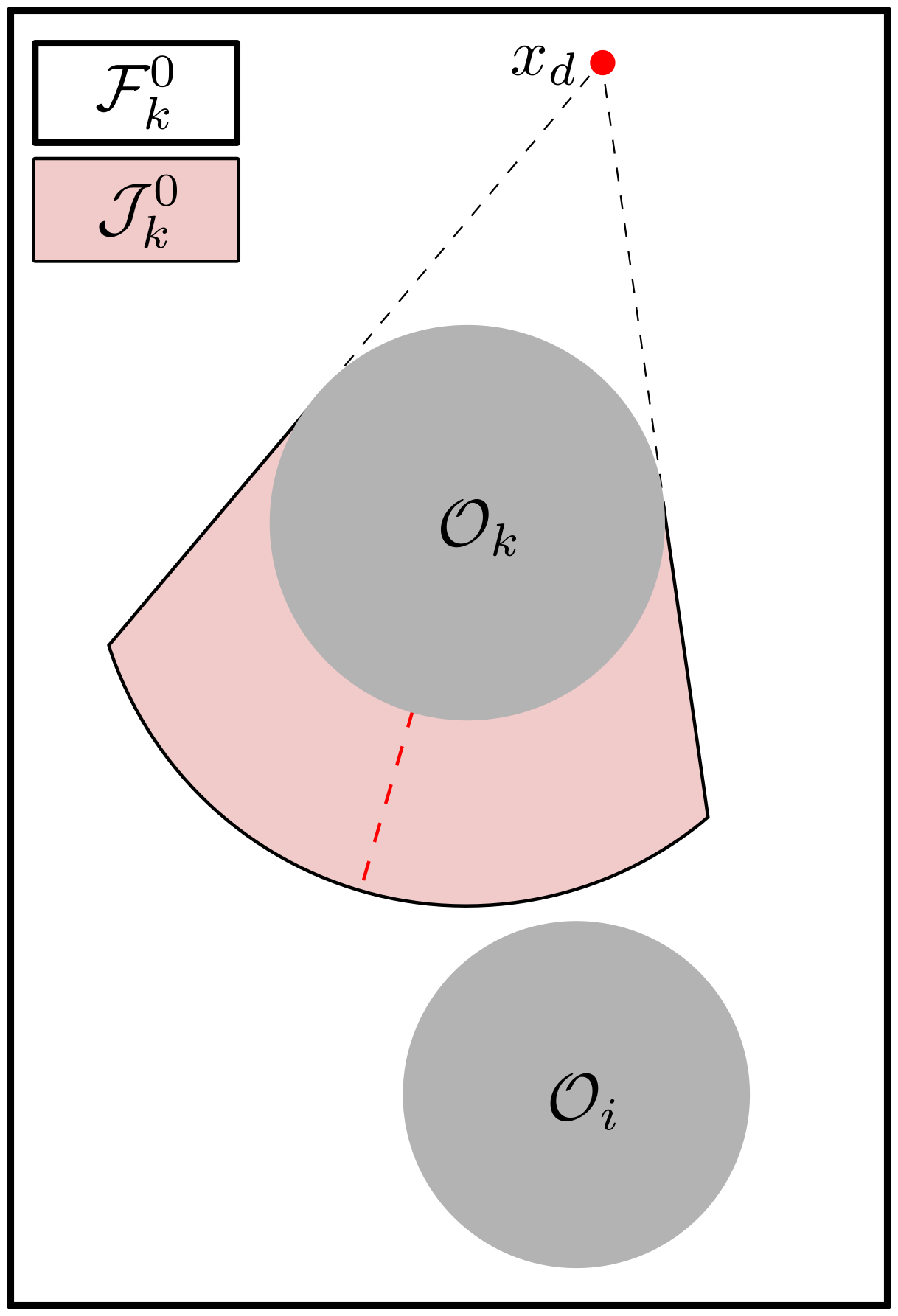}
\caption{2D illustration of the flow and jump sets for the \textit{motion-to-destination} mode associated with obstacle $\mathcal{O}_k,\;k\in\mathbb{I}$.}
\label{fig3}
\end{figure}

In this mode, the robot moves straight to the target under the nominal control $u_d(x)$. Considering obstacle $k\in\mathbb{I}$, the flow and jump sets associated with obstacle $\mathcal{O}_k$, depicted in Fig. \ref{fig3}, are defined as follows:
\begin{align}\label{sets_m0}
\mathcal{F}_k^0&:=\overline{\mathcal{X}\setminus\mathcal{A}_k(x_d)},\;\mathcal{J}_k^0:=\mathcal{A}_k(x_d),
\end{align}
where $\mathcal{A}_k(x_d)$ is the {\it active region} defined in \eqref{active_region}. The flow and jump sets for the mode \( m=0 \), considering all obstacles, are defined as:
\begin{align}\label{set_0}
    \mathcal{F}_0 := \Tilde{\mathcal{F}}_0 \times \mathbb{I}, \quad \mathcal{J}_0 := \Tilde{\mathcal{J}}_0 \times \mathbb{I},
\end{align}
where the \textit{motion-to-destination} mode is selected at each position \( x \) within the intersection of the flow sets, \( \Tilde{\mathcal{F}}_0 := \cap_{k \in \mathbb{I}} \mathcal{F}_k^0 \), for all obstacle indices \( k \in \mathbb{I} \). Additionally, at each position \( x \) in the union of the jump sets \( \Tilde{\mathcal{J}}_0 := \cup_{k\in\mathbb{I}}\mathcal{J}_k^0 = \mathcal{V}(x_d) \), where $\mathcal{V}(x_d)$ is the {\it active free space} defined in \eqref{active_freespace}, a jump to the \textit{obstacle-avoidance} mode can occur for any obstacle index \( k \in \mathbb{I} \).
\subsubsection{Obstacle-avoidance mode ($m=\pm1$)}\label{projection_mode_design}
\begin{figure}[h!]
\centering
\includegraphics[width=0.7\columnwidth]{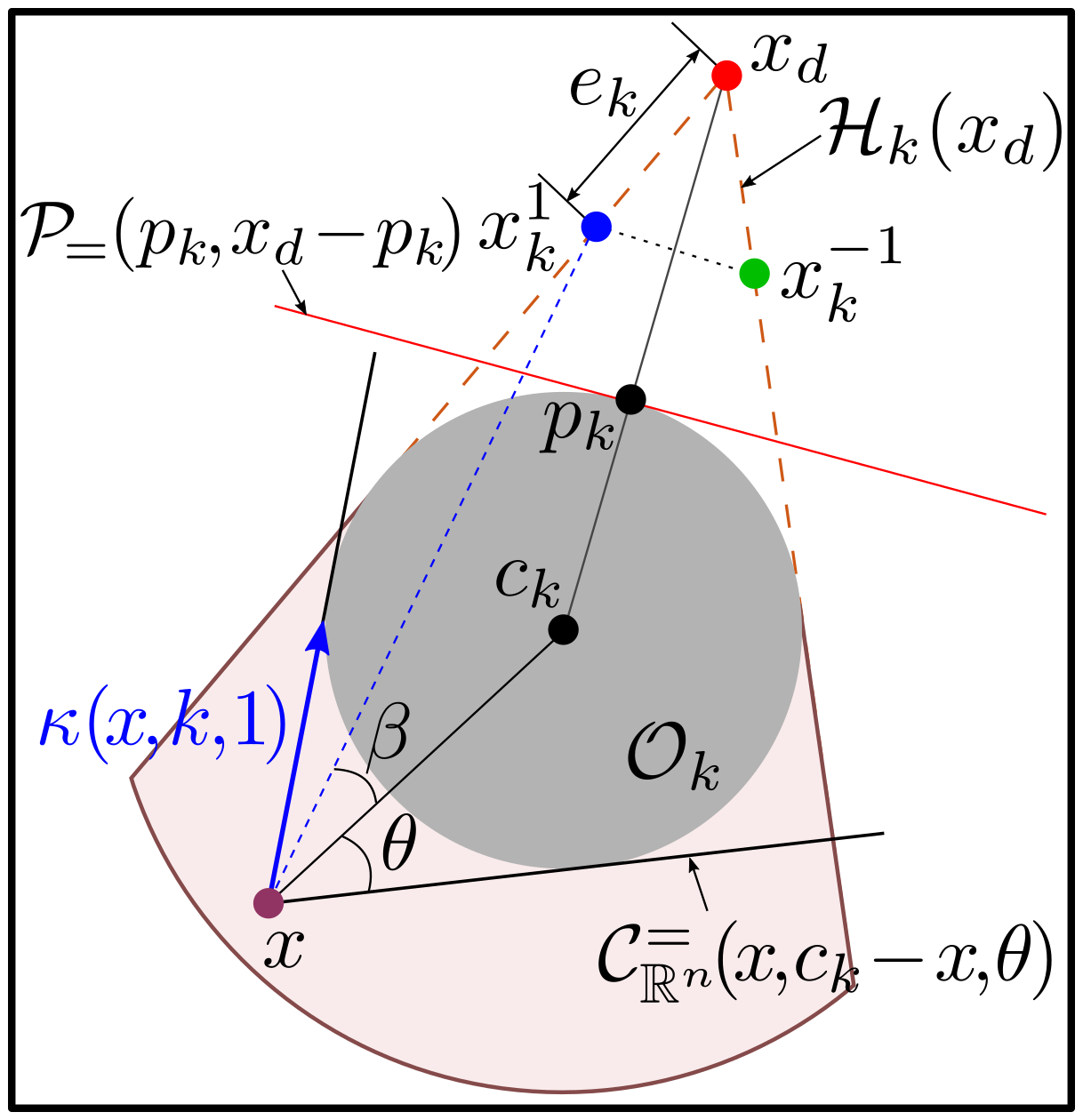}
\caption{Construction of the control in the {\it obstacle-avoidance} mode for a 2D case.}
\label{fig_u_m}
\end{figure}

\begin{figure}[h!]
\centering
\includegraphics[width=\columnwidth]{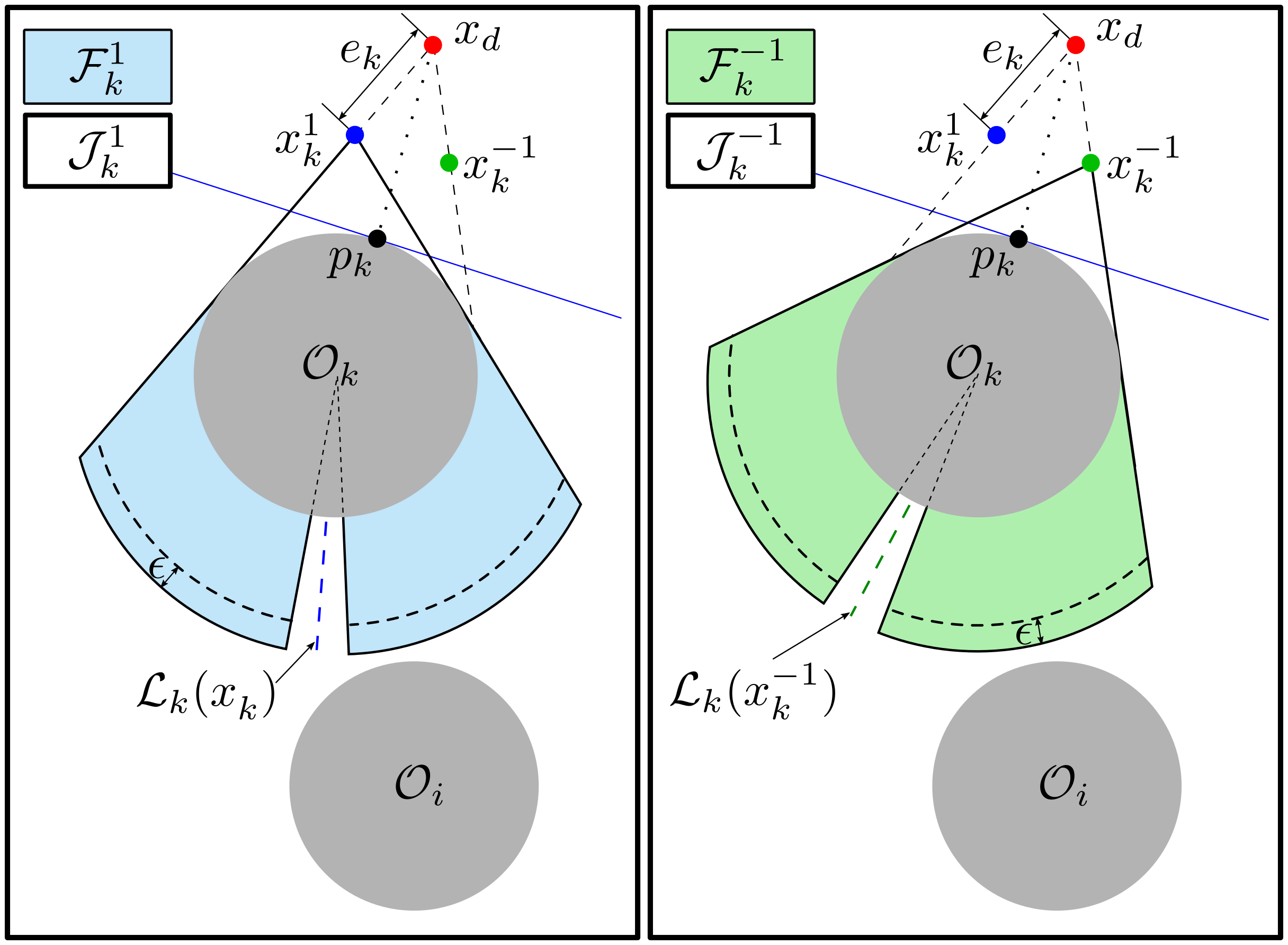}
\caption{2D illustration of the flow and jump sets for the \textit{obstacle-avoidance} mode associated with obstacle $\mathcal{O}_k,\;k\in\mathbb{I}$.}
\label{fig4}
\end{figure}

In this mode of operation, the robot will engage in a local optimal obstacle avoidance maneuver. To this end, we consider two virtual destinations, $x_k^{1}$ and $x_k^{-1}$, which are designed to be on the {\it hat} $\mathcal{H}_k(x_d)$, defined in \eqref{hat}, of obstacle $\mathcal{O}_k$ and symmetrical with respect to the hat axis $(c_k-x_d)$, as shown in Fig. \ref{fig_u_m}. The introduction of two virtual destinations is motivated by the observation that each virtual destination generates a distinct set of undesired equilibria. By appropriately selecting the virtual destination during the avoidance maneuver (via hybrid feedback), the vehicle is prevented from becoming trapped at these undesired equilibria, see \cite{SoulaimaneHybTr}. Moreover, by projecting the nominal control $\bar{\kappa}(x,k,m):=\gamma(x_k^m-x)$ onto the surface of the cone, with vertex at \( x \), enclosing obstacle \( \mathcal{O}_k \), we aim to avoid the obstacle with minimal deviation from the nominal direction, as illustrated in Fig. \ref{fig_u_m}, thereby generating optimal obstacle maneuvers, see also \cite{Ishak2023}. These virtual destinations are chosen as follows: 
\begin{subequations}\label{virtual_destinations}
\begin{align}
    x_k^1\in\mathcal{H}_k(x_d)\cap\mathcal{P}_{\geq}(p_k,x_d-p_k), \label{virtual_1}\\
    x_k^{-1}=x_d-\pi_r\left(\frac{c_k-x_d}{\|c_k-x_d\|}\right)(x_k^1-x_d)
\end{align}
\end{subequations}
where $p_k:=c_k+r_k\frac{x_d-c_k}{\|x_d-c_k\|}$ and $\|x_d-x_k^m\|=:e_k>0$. Note that the choice of $x_k^1$ is not unique and can be any point on the hat $\mathcal{H}_k(x_d)$ of the enclosing cone within the half-space $\mathcal{P}_{\geq}(p_k,x_d-p_k)$, and satisfying $\|x_d-x_k^m\|=e_k$.  The term $\kappa(x,k,m)$ used in \eqref{hyb_ctrl_L}, which is the control in the {\it obstacle-avoidance mode} (\textit{i.e.,} $m=\pm1$), is given as follows:
\begin{align}\label{ctrl_m11}
    \kappa(x,k,m)=\bar{\kappa}(x,k,m)-\tau(x,k,m)\frac{c_k-x}{\|c_k-x\|},
\end{align}
where $\tau(x,k,m)=\|\bar{\kappa}(x,k,m)\|\sin(\theta(x,k)-\beta(x,k,m))/ \sin(\theta(x,k))$, $\beta(x,k,m)=\angle(c_k-x,\bar{\kappa}(x,k,m))$, and $\theta(x,k)=\arcsin(r_k/\|x-c_k\|)\in(0,\pi/2]$ with $k\in\mathbb{I}$ and $m\in\{-1,1\}$. The control law \eqref{ctrl_m11} is a scaled parallel projection of the nominal controller $\bar{\kappa}(x,k,m)$, with respect to the virtual destination $x_k^m$, onto the line tangent to obstacle $\mathcal{O}_k$, ensuring a minimal angle with $\bar{\kappa}(x,k,m)$ and the continuity of $\kappa(x,k,m)$ at the {\it exit set}. The optimization problem with solution $\kappa(x,k,m)$ is given in the following lemma.
\begin{lemma}\label{Opt_lem}
    Consider obstacle $\mathcal{O}_k$, a virtual destination $x_k^m$ and the {\it active region} $\mathcal{A}_k(x_k^m)$ where $k\in\mathbb{I}$ and $m\in\{-1,1\}$. For each $(x,k,m)$, the control law $\kappa(x,k,m)$, given in \eqref{ctrl_m11}, is the unique solution of the optimization problem given by
    \begin{subequations}\label{Opt_pb}
     \small
      \begin{align}
            &\min_{v}\angle(x_k^m-x,v),\label{constr1}\\
            \textrm{s.t.}\;&v\in\mathcal{V}(c_k-x,\theta(x,k)),~&&\textrm{if}\;x\in\mathcal{A}_k(x_k^m),\\
            &v=\bar{\kappa}(x,k,m),~&&\textrm{if}\;x\in\mathcal{E}_k(x_k^m),\label{constr2}
    \end{align} 
    
    \end{subequations}
    where $\mathcal{A}_k(x_k^m)$ and $\mathcal{E}_k(x_k^m)$ are defined in \eqref{active_region} and \eqref{12} respectively.
\end{lemma}
The proof of Lemma \ref{Opt_lem} is the same as in \cite[Appendix A]{Ishak2023}, with the virtual destination $x_k^m$ replacing the actual destination $x_d$ and the {\it active region} $\mathcal{A}_k(x_k^m)$ replacing the {\it shadow region} $\mathcal{S}_k(x_k^m)$. Note that the constraint \eqref{constr2} ensures the uniqueness of the solution to the optimization problem \eqref{Opt_pb} and the continuity of $\kappa(x,k,m)$ at the {\it exit set} $\mathcal{E}_k(x_k^m)$. The control law \eqref{ctrl_m11} steers the robot tangentially to the obstacle, providing a more efficient path compared to approaches like the bug algorithms, which initiate obstacle avoidance only upon contact or in close proximity to the obstacle, as seen in hybrid approaches such as \cite{SoulaimaneHybTr,Mayur2022,Mayur_non_convex}. The equilibrium points generated by the control input \eqref{ctrl_m11} in the {\it obstacle-avoidance} mode can be obtained by setting $\kappa(x,k,m)=0$ for $x\in\mathcal{A}_k(x_k^m)$, $m\in\{-1,1\}$, and $k\in\mathbb{I}$. It is clear that  $\kappa(x,k,m)=0$ holds only if $\beta(x,k,m)=0$. The set of equilibria is then the line passing through the center $c_k$ and the virtual destination $x_k^m$ limited by the {\it active region} $\mathcal{A}_k(x_k^m)$. It is given by $\mathcal{L}_k(x_d^m):=\mathcal{L}(x_k^m,c_k)\cap\mathcal{A}_k(x_k^m)$.\\
The flow and jump sets of the \textit{obstacle-avoidance} mode, considering obstacle $\mathcal{O}_k$, $k\in\mathbb{I}$, are illustrated in Fig. \ref{fig4} and defined as follows:
\begin{align}\label{sets_11}
    \mathcal{F}_k^m&:=\mathcal{A}_k(x_k^{m})\setminus\mathcal{C}_{\mathcal{X}}^{<}(c_k,v_k^m,\varphi_k^m),\mathcal{J}_k^m:=\overline{\mathcal{X}\setminus\mathcal{F}_k^m},
\end{align}
where $v_k^m:=c_k-x_k^m$.
To ensure that the jump from the {\it motion-to-target} mode to the {\it obstacle-avoidance} mode is valid everywhere in $\mathcal{J}_0$ ({\it i.e., $\forall(x,k)\in\mathcal{J}_0,~ M(x,k,0)\neq\varnothing$}), we design the angles $\varphi_k^1,\varphi_k^{-1}$ in \eqref{sets_11} as per Lemma \ref{lem1} as follows:
\begin{equation}\label{phi}
\small
    \varphi_k^1=\varphi_k^{-1}=\varphi<\min\left\{\frac{\angle{(v_k^1,v_k^{-1})}}{2},\frac{\pi-\angle{(v_k^1,v_k^{-1}})}{2}\right\}.
\end{equation}
Note that a conic subset is subtracted from the {\it active region} for the modes $m=\pm1$ in \eqref{sets_11}, which excludes the equilibria from the flow set, leaving the system with a unique equilibrium point at the target location for the mode $m=0$. The flow and jump sets for mode $m\in\{-1,1\}$ and considering all the obstacles are defined as follows:
\begin{align}\label{set_m}    \mathcal{F}_m:=\bigcup\limits_{k\in\mathbb{I}}\left(\mathcal{F}_k^m\times\{k\}\right),\;\mathcal{J}_m:=\bigcup\limits_{k\in\mathbb{I}}\left(\mathcal{J}_k^m\times\{k\}\right). 
\end{align}
\subsubsection{Operation mode switching scheme}\label{Switching_scheme}

The jump maps are designed to effectively switch between the operation modes to attain the following objectives:
\begin{itemize}
    \item Avoid every encountered obstacle through local optimal maneuvers.
    \item Avoid obstacles one by one to ensure safety and offer the possibility of sensor-based implementation.
    \item Converge to the destination $x_d$ from any initial position in the obstacle-free space.
\end{itemize}
We first define the smoothing functions as follows:
\begin{align}
    &\mu(x,k,m):=\left(1+\frac{e_k}{\|x-x_k^m\|}\frac{\beta(x,k,m)}{\theta(x,k)}\right),\label{smoothing_fct1}\\
    &\alpha(x,k):=\begin{cases}
        1 & d(x,\mathcal{O}_k)<\Bar{r}_k-\epsilon,\\
        \frac{\Bar{r}_k-d(x,\mathcal{O}_k)}{\epsilon} & \Bar{r}_k-\epsilon\leq d(x,\mathcal{O}_k)\leq\Bar{r}_k,\\
        0 & d(x,\mathcal{O}_k)>\Bar{r}_k,
    \end{cases}\label{smoothing_fct2}
\end{align}
with $0<\epsilon\leq\Bar{\epsilon}$ and $\Bar{\epsilon}:=\min\limits_{k\in\mathbb{I}}\Bar{r}_k$. In fact, the scalar function $\alpha(x,k)$ ensures a smooth transition from the \textit{motion-to-destination} mode to the \textit{obstacle-avoidance} mode. The scalar function $\mu(x,k,m)$, together with the modified switching scheme of the mode $m$ that will be designed in Subsection \ref{continuity_optimality}, ensures a smooth transition from the \textit{obstacle-avoidance} mode to the \textit{motion-to-destination} mode.\\
Now, we define the jump map $K(\cdot)$ of the obstacle selector as
\begin{align}\label{jump_map_K}
    K(x,k,m):=\begin{cases}
     k  &x\in \mathcal{J}_k^m,\;m\in\{-1,1\},\\
     k' &x\in {\mathcal{J}_{k'}^0},\; m=0,
      \end{cases}
\end{align}
and the jump map $M(\cdot)$ of the mode selector as
\begin{align}\label{jump_map_M}
    M(x,k,m):=\begin{cases}
    0 &x\in \mathcal{J}_k^m,\;m\in\{-1,1\},\\
    B(x,k) &x\in {\mathcal{J}_{k}^0},\; m=0,
      \end{cases}
\end{align}
where $B(\cdot)$ is defined as:
\begin{align}\label{jump_map_B}
    B(x,k):=\begin{cases}
    1&x\in\mathcal{C}_k^1,\\
    -1 &x\in\mathcal{C}_k^{-1}\\
    \{-1,1\}&x\in\mathcal{C}_k ,
      \end{cases}
\end{align}
with $\mathcal{C}_k^1=\mathcal{C}_{\mathbb{R}^n}^{\geq}(c,v_k^1,\varphi_k^1)\setminus\mathcal{C}_{\mathbb{R}^n}^{\geq}(c,v_k^{-1},\varphi_k^{-1})$, $\mathcal{C}_k^{-1}=\mathcal{C}_{\mathbb{R}^n}^{\geq}(c,v_k^{-1},\varphi_k^{-1})\setminus\mathcal{C}_{\mathbb{R}^n}^{\geq}(c,v_k^1,\varphi_k^1)$, and $\mathcal{C}_k=\mathcal{C}_{\mathbb{R}^n}^{\geq}(c,v_k^1,\varphi_k^1)\cap\mathcal{C}_{\mathbb{R}^n}^{\geq}(c,v_k^{-1},\varphi_k^{-1})$.
Note that the construction of the flow and jump sets \eqref{sets_11} of the {\it obstacle-avoidance} mode are such that the set of undesired equilibria belongs to the jump set of the corresponding {\it obstacle-avoidance} mode ($m=\pm1$). 
\subsection{Safety and stability analysis}\label{subsection:stability}
In this subsection, we establish the safety and stability properties of our hybrid closed-loop system. To this end, we define the augmented state vector as
\begin{align}
\xi :=(x,k,m)\in\mathbb{R}^n\times\mathbb{I}\times\mathbb{M},
\end{align}
and the overall flow and jump maps as
\begin{align}
    \xi\mapsto \mathrm{F}(\xi):=(u(\xi),0,0),\label{flow_map}\\
    \xi\mapsto \mathrm{J}(\xi):=(x,K(\xi),M(\xi)).\label{jump_map}
\end{align}
Then, the resulting hybrid closed-loop system can be written as
\begin{align}\label{Cl_ctrl}
    \begin{cases}
    \Dot{\xi}=\mathrm{F}(\xi)
    &\xi\in\mathcal{F},\\
    \xi^{+}\in \mathrm{J}(\xi)&\xi\in\mathcal{J},
    \end{cases}
\end{align}
and its representation with the hybrid data is given by $\mathcal{H}:=(\mathcal{F},\mathrm{F},\mathcal{J},\mathrm{J})$. We also define the augmented free space and the desired equilibrium set as follows: 
\begin{align}\label{Hyb_freespace_att}
    \mathcal{K}:=\mathcal{X}\times\mathbb{I}\times\mathbb{M},\;
    \mathcal{A}:=\{x_d\}\times\mathbb{I}\times\mathbb{M}.
\end{align}
To analyze our hybrid closed-loop system, we first establish its well-posedness by showing that it complies with the hybrid basic conditions \cite[Assumption 6.5]{Sanfelice}, as shown in the next lemma.
\begin{lemma}\label{lem2}
   The hybrid closed-loop system \eqref{Cl_ctrl} represented by its data $\mathcal{H}$, satisfies the following hybrid basic conditions:
   \begin{itemize}
       \item [i)] The flow set $\mathcal{F}$ and the jump set $\mathcal{J}$, defined in \eqref{sets}, are closed subsets of $\mathcal{K}$.
       \item [ii)] The flow map $\mathrm{F}$, defined in \eqref{flow_map}, is outer semicontinuous
and locally bounded relative to $\mathcal{F}$, $\mathcal{F} \subset \text{dom} \mathrm(F)$, and
$\mathrm{F}(\xi)$ is convex for every $\xi\in\mathcal{F}$.
\item [iii)] The jump map $\mathrm{J}$, defined in \eqref{jump_map}, is outer semicontinuous
and locally bounded relative to $\mathcal{J}$, and $\mathcal{J} \subset \text{dom} \mathrm(J)$.
   \end{itemize}
\end{lemma}
\begin{proof}
See Appendix \ref{appendix:Lem2}.
\end{proof}

\begin{figure}[h!]
\centering
\includegraphics[width=0.7\columnwidth]{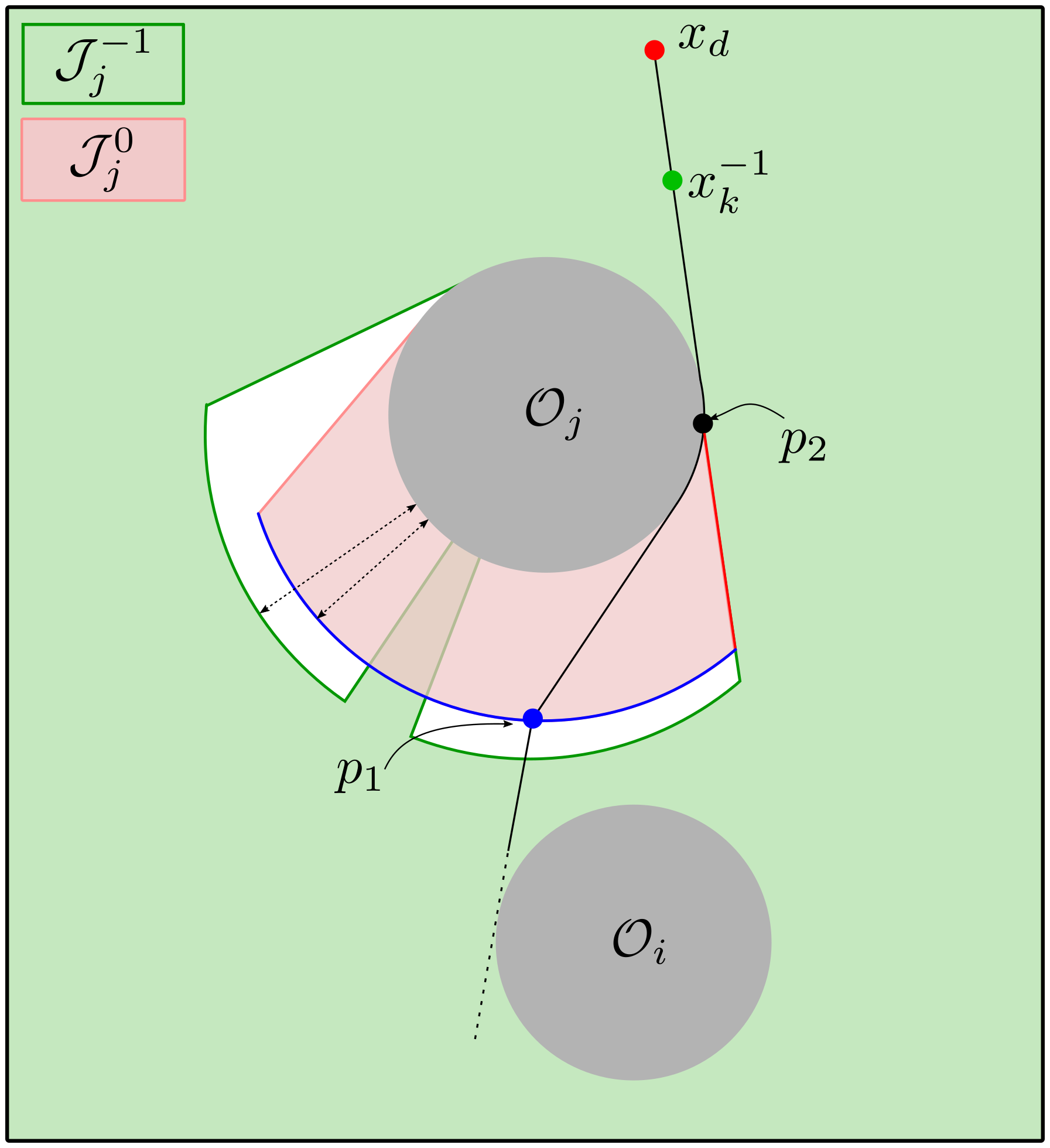}
\caption{Navigation under the Zeno-free switching scheme. The robot starts in the {\it motion-to-destination} mode and proceeds until it reaches the bottom boundary of the jump set $\mathcal{J}^0_j$ (blue arc) at point $p_1$. At this point, the obstacle index $k$ switches to $j$, and the mode changes to {\it obstacle-avoidance}. Since the blue boundary lies outside the jump set of the {\it obstacle-avoidance} mode, the robot cannot switch back immediately to {\it motion-to-destination}. When it reaches the exit set (red segment)—the common boundary of $\mathcal{J}^{-1}_j$ and $\mathcal{J}^0_j$—at point $p_2$, the mode switches back to {\it motion-to-destination} while the index selector remains unchanged. This prevents reactivation of the {\it obstacle-avoidance} mode, in accordance with the switching scheme \eqref{jump_map_M_mod}.
}
\label{fig_jump_mod_rev1}
\end{figure}

Robot navigation is said to be safe if the state $x$ evolves in the obstacle-free space $\mathcal{X}$ at all times. For the hybrid closed-loop system \eqref{Cl_ctrl}, this is equivalent to showing forward invariance \cite{Forward_invariance_sanfelice} of the augmented obstacle-free space $\mathcal{K}$. The following theorem will present our results concerning safe and global navigation under the proposed hybrid controller.
\begin{theorem}\label{the1}
    Consider the augmented free space $\mathcal{K}$, described in \eqref{Hyb_freespace_att}, and the hybrid closed-loop system \eqref{Cl_ctrl}. Then, the following statements hold:
    \begin{itemize}
        \item [i)] The augmented free space $\mathcal{K}$ is forward invariant.
        \item [ii)] All Zeno-free solutions converge to the set $\mathcal{A}$.
    \end{itemize}
\end{theorem}
\begin{proof}
See Appendix \ref{appendix:the1}.
\end{proof}
Theorem \ref{the1} states that the robot safely navigates the free space and reaches the destination for all Zeno-free solutions. In practice, Zeno-free behaviors are enforced for any initial condition in the augmented free space $\mathcal{K}$ by prioritizing flows over jumps, thereby ensuring global attractivity of the target location. To guarantee this property by construction, we introduce a modified version of the jump map:
\begin{align}\label{jump_map_M_mod}
    \hat{M}(x,k,m):=
    \begin{cases}
        (k,0) & x \in \mathcal{J}_k^m,\, m\in\{-1,1\},\\
        (k',B(x,k')) & x \in \mathcal{J}_{k'}^0,\, m=0,\, k\neq k',\\
        (k,m) & x \in \mathcal{J}_k^0,\, m=0.
    \end{cases}
\end{align}
This switching scheme allows transitions from \textit{obstacle-avoidance} to \textit{motion-to-destination} mode only when the obstacle index jumps to a new value on $\mathcal{J}_k^0$, preventing infinite switching where $\mathcal{J}_k^0$ and $\mathcal{J}_k^{\pm1}$ coincide (see Fig. \ref{fig_jump_mod_rev1}). Moreover, by selecting larger ranges of {\it active regions} for virtual destinations than for the target, the mode cannot switch back to \textit{motion-to-destination} immediately after entering \textit{obstacle-avoidance} through the bottom boundary of $\mathcal{J}_k^0$ ({\it i.e.,} $\partial\mathcal{J}_k^0\cap\partial\mathcal{B}(c_k,\bar{r}_k)$), as illustrated in Fig. \ref{fig_jump_mod_rev1}. This design ensures Zeno-free jumps by construction. However, to guarantee safety, the initial obstacle index $k(0,0)$ must differ from the first obstacle to be avoided; otherwise, the robot would not switch to \textit{obstacle-avoidance} when entering that obstacle’s {\it active region}, risking collision. The initialization condition is given by $k(0,0)=\arg\min_{j\in\mathbb{K}}d(x,\mathcal{O}_j)$, where $\mathbb{K}:=\{j\in\mathbb{I}|\mathcal{L}_s(x(0,0),x_d)\cap\mathcal{O}_j\neq 0\}.$ For the analysis, the original switching scheme is adopted, since it offers a simpler formulation with implementation performance identical to the modified version.\\

In the next subsection, we demonstrate that our control strategy ensures the continuity of the control input \eqref{hyb_ctrl_L} and the optimality of the local obstacle avoidance maneuvers.
 \subsection{Continuity and optimality}\label{continuity_optimality}
\begin{figure}[h!]
     \centering
     \subfloat[]{\includegraphics[width=0.3\linewidth,keepaspectratio]{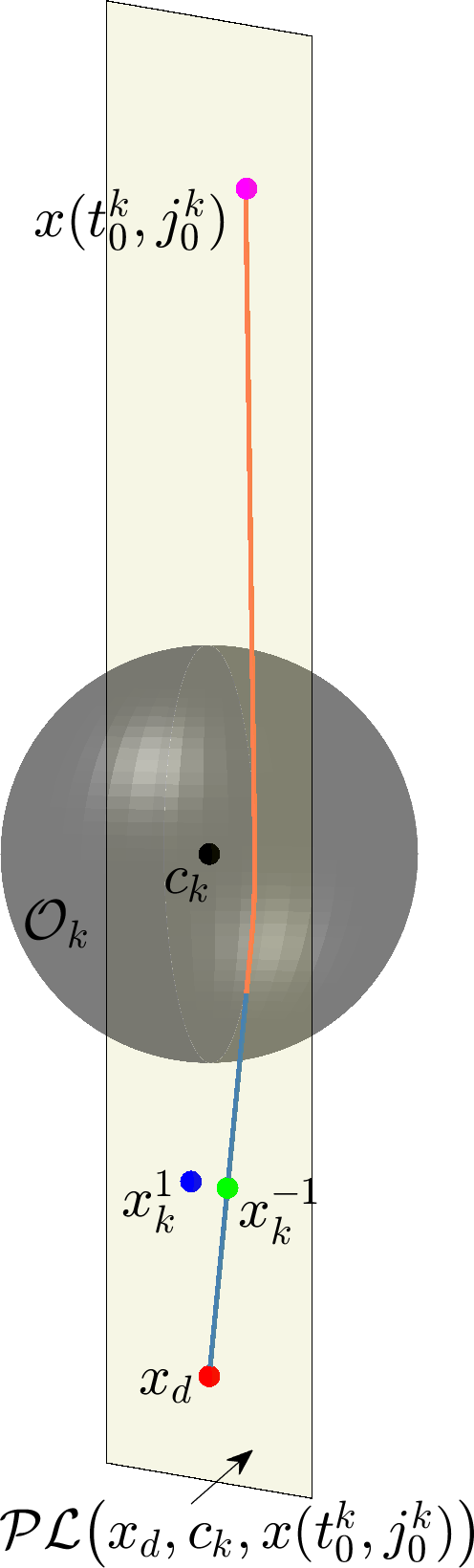}\label{fig:fig5_left}}
    \hfill
     \subfloat[]{\includegraphics[width=0.3\linewidth,keepaspectratio]{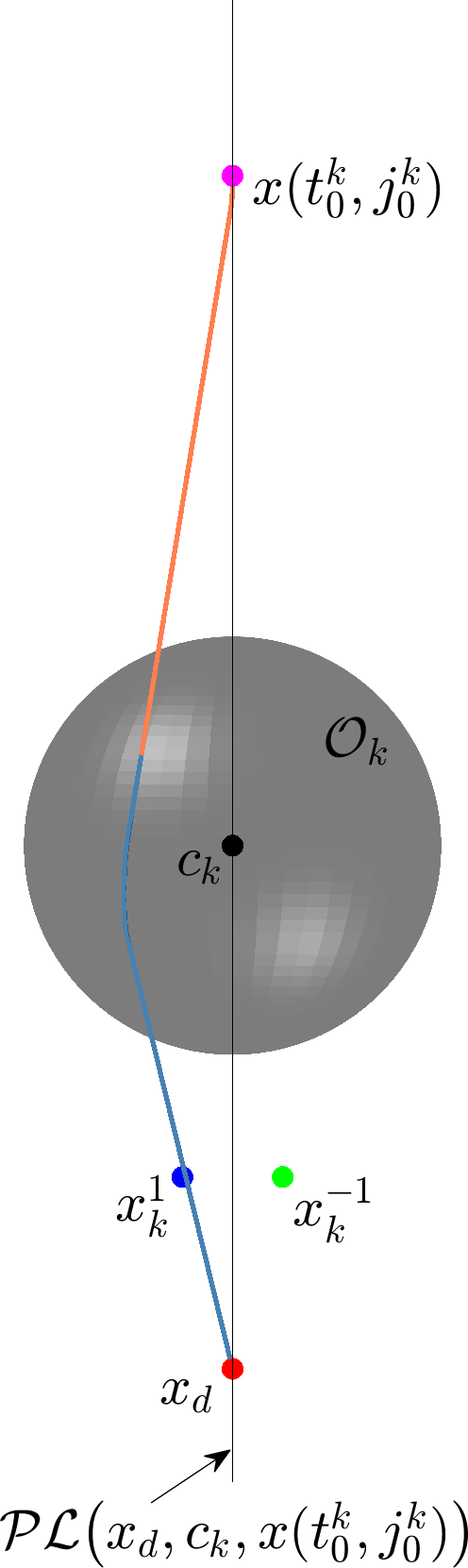}\label{fig:fig5_middle}}
      \hfill
     \subfloat[]{\includegraphics[width=0.3\linewidth,keepaspectratio]{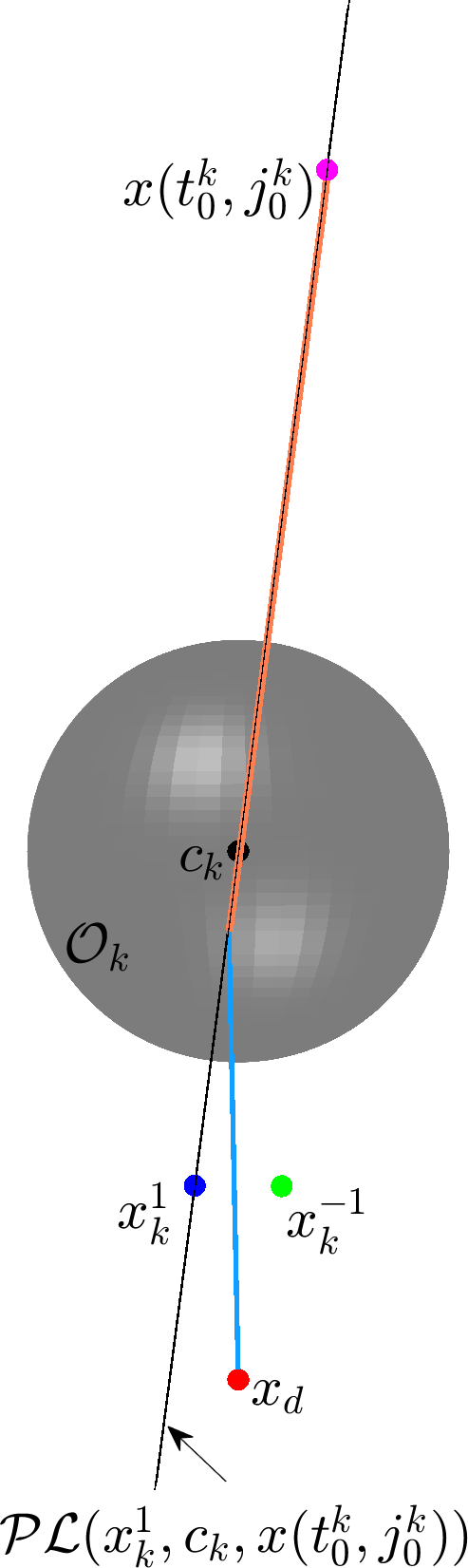}\label{fig:fig5_right}}\\
     \caption{Illustration of the property in Lemma \ref{lem6} in a 3D space. In the left figure (a), the virtual destinations $x_k^{\pm1}$ belong to the plane $\mathcal{PL}(x_d,c_k,x(t_0^k,j_0^k))$, resulting in a trajectory belonging to the same plane for the obstacle $\mathcal{O}_k$, where the {\it obstacle-avoidance} mode is represented by the orange curve and the {\it motion-to-destination} mode is represented by the blue curve. In the middle figure (b), the virtual destinations $x_k^{\pm1}$ do not belong to the plane $\mathcal{PL}(x_d,c_k,x(t_0^k,j_0^k))$, resulting in a trajectory that does not belong to the plane $\mathcal{PL}(x_d,c_k,x(t_0^k,j_0^k))$. The right figure (c) shows that the trajectory generated in figure (b) does not belong to a single plane but to two different planes, as the virtual destinations are not on the plane $\mathcal{PL}(x_d,c_k,x(t_0^k,j_0^k))$.}
     \label{fig:Lemma5_v}
\end{figure}

\begin{figure}[h!]
\centering
\subfloat[]{
\includegraphics[width=0.48\linewidth,keepaspectratio]{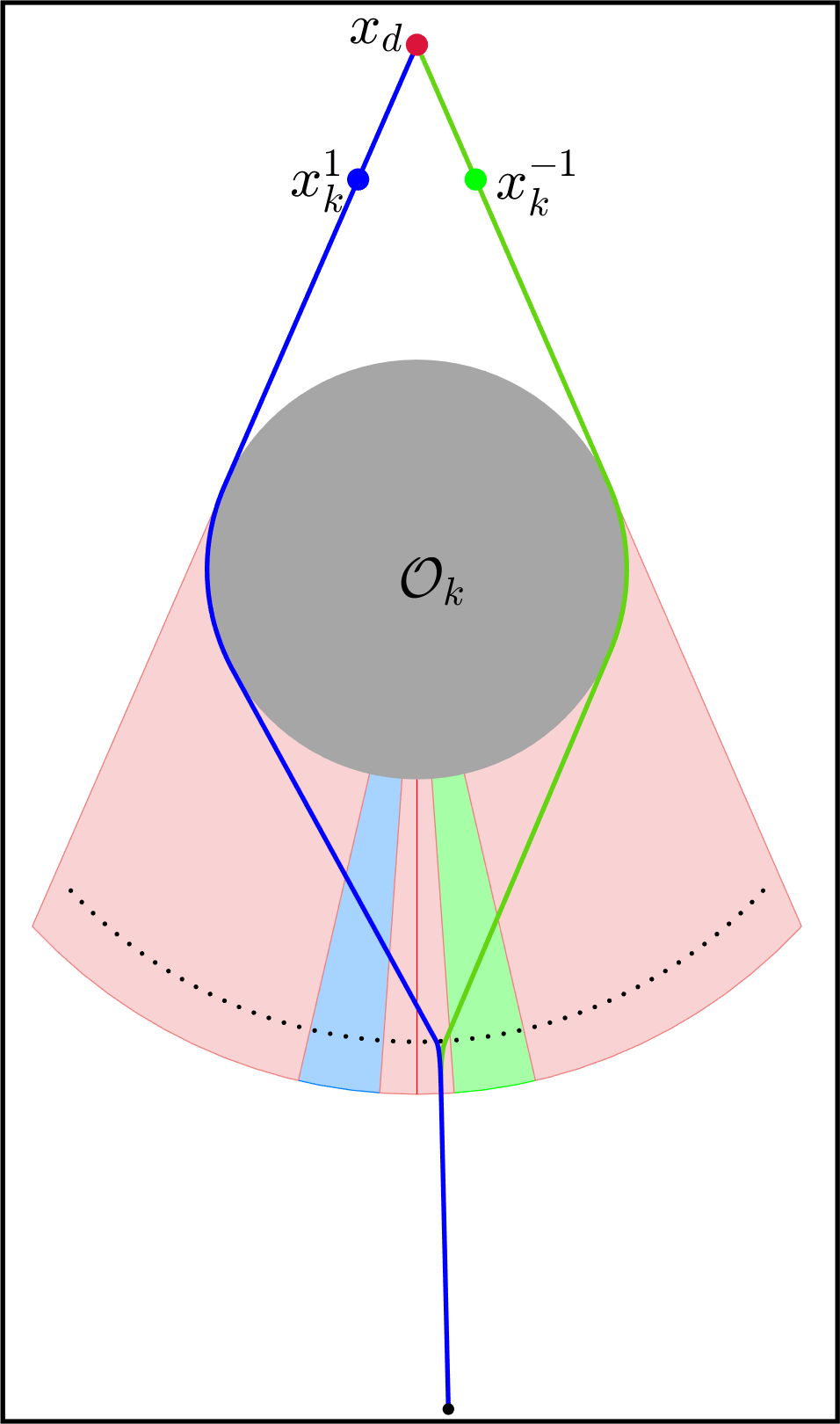}\label{fig_prop1}}
\subfloat[]{
\includegraphics[width=0.48\linewidth,keepaspectratio]{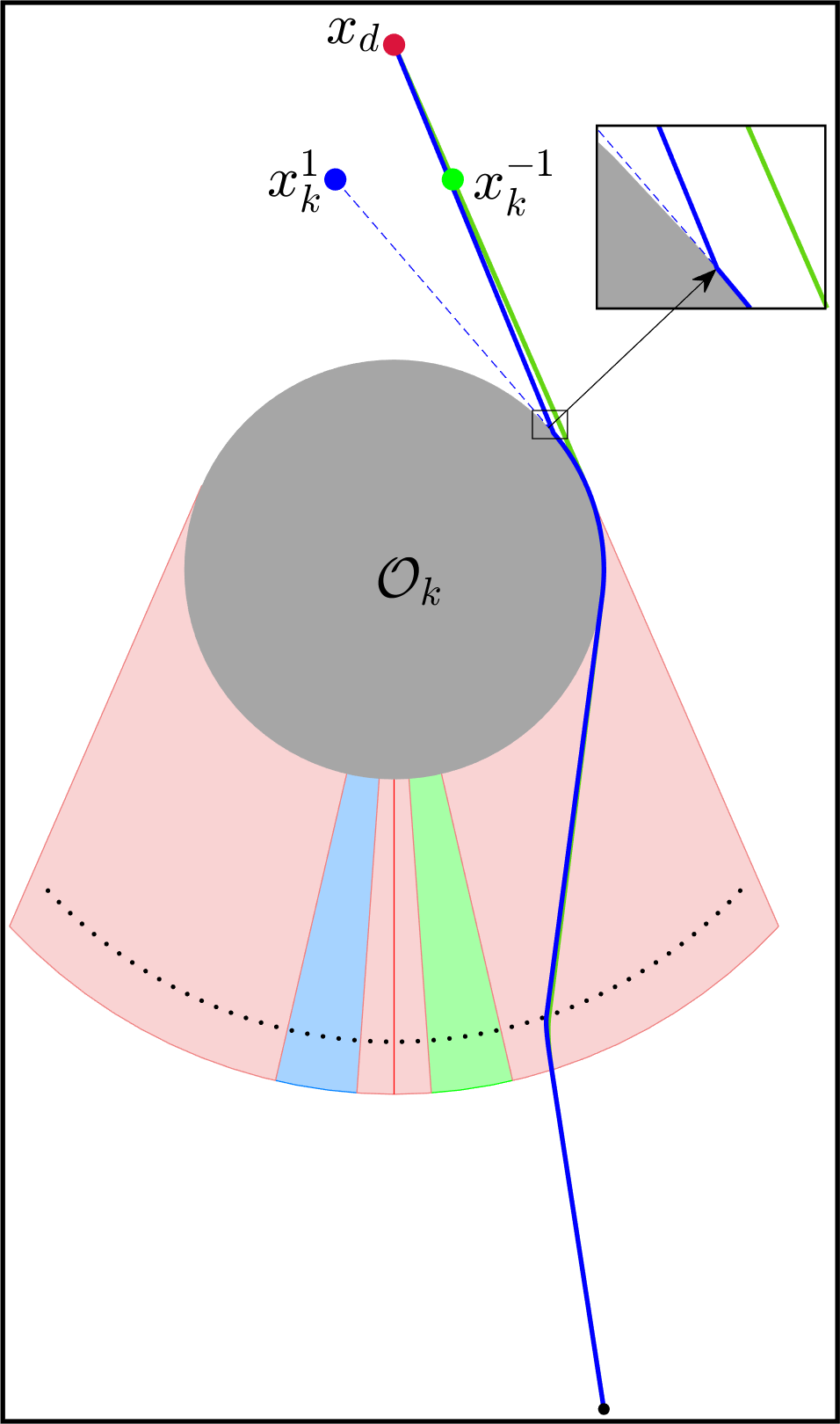}\label{fig_prop2}}
\caption{
Effects of selecting the navigation mode according to Proposition \ref{proposition1}. In the left figure (a), when first entering the jump set $\mathcal{J}_k^0$ through the hysteresis region (pink region), the mode $m$ switches to {\it obstacle-avoidance} mode where the blue trajectory is generated by selecting the virtual destination $x_k^1$ and the green trajectory is generated by selecting the closest virtual destination $x_k^{-1}$ to the robot's position. Both trajectories are smooth, but the blue trajectory is longer than the green trajectory. In the right figure (b), the blue trajectory is longer and non-smooth. The green (blue resp.) region in the jump set $\mathcal{J}_k^0$ is where, if the robot operates in the {\it motion-to-destination} mode, the mode $m$ can only jump to  $m=-1$ ($m=1$ resp.).}
\label{figProposition}
\end{figure}

 To ensure continuity of the control input and optimality of the avoidance maneuvers when implementing the hybrid control \eqref{hyb_ctrl}, two properties of the proposed hybrid system will be utilized. In fact, the closed-loop hybrid system defined in \eqref{Cl_ctrl} offers flexibility in choosing the virtual destinations (see \eqref{virtual_1}) as well as the operation mode $m$ (set-valued map in \eqref{jump_map_B}). In this section, we show that, for a given obstacle \( k \in \mathbb{I} \), the virtual destinations in \eqref{virtual_destinations} can be selected to guarantee a two-dimensional motion. Moreover, when the robot is in the hysteresis region ({\it i.e.,} $x\in \mathcal{J}_k^0\cap \mathcal{C}_k$), the mode $m$ can be forced in \eqref{jump_map_B} to the value that ensures the virtual destination $x_k^m$ is the closest to the robot. In the following, we present the detailed design process that guarantees continuity of the control input and local optimality of the obstacle avoidance maneuver. The proposed control strategy generates planar trajectories during the time interval in which a given obstacle $\mathcal{O}_k$ is being avoided, as stated in the following lemma: 
\begin{lemma}\label{lem6}
Let the plane  spanned by two non-colinear vectors $(q_1-y) \in\mathbb{R}^n$ and $(q_2-y)\in\mathbb{R}^n$ be denoted by $\mathcal{PL}(q_1,q_2,y)$.
Consider the closed-loop hybrid system \eqref{Cl_ctrl}. For a given obstacle index $k\in\mathbb{I}$, if the virtual destinations $x_k^{-1}$ and $x_k^1$ belong to the plane $\mathcal{PL}(x_d,c_k,x(t_0^k,j_0^k))$ when the destination $x_d$, the obstacle's center $c_k$ and the position $x(t_0^k,j_0^k)$ are not aligned, the motion takes place in the plane $\mathcal{PL}(x_d,c_k,x(t_0^k,j_0^k))$ where $(t_0^k, j_0^k)$ is the hybrid time at which obstacle $k$ is selected. If the points $x_d$, $c_k$ and $x(t_0^k,j_0^k)$ are aligned, the motion takes place in the plane $\mathcal{PL}(x_d,c_k,y)$ for a given $y\in\mathbb{R}^n\setminus\mathcal{L}(x_d,c_k)$ such that $x_k^m\in\mathcal{PL}(x_d,c_k,y)$ and $m\in\{-1,1\}$. 
\end{lemma}
\begin{proof}
See Appendix \ref{appendix:Lem6}.
\end{proof}
The result of Lemma \ref{lem6} requires selecting the virtual destinations associated with obstacle $k\in\mathbb{I}$ depending on the robot's position when it first enters the jump set $\mathcal{J}_0^k$ of obstacle $k$. Suppose the virtual destinations, defined in \eqref{virtual_destinations}, belong to the two-dimensional plane $\mathcal{PL}(x_d,c_k,x(t_0^k,j_0^k))$  when $x(t_0^k,j_0^k)\in\mathcal{J}_k^0\setminus\mathcal{L}(x_d,c_k)$. In this case, the motion generated by the closed-loop system \eqref{Cl_ctrl}, while obstacle $k$ is selected, is two-dimensional and takes place on the plane $\mathcal{PL}(x_d,c_k,x(t_0^k,j_0^k))$. It is clear that in the case where $x(t_0^k,j_0^k)\in\mathcal{J}_k^0\cap\mathcal{L}(x_d,c_k)$, the plane of motion can be any plane $\mathcal{PL}(x_d,c_k,y)$ such that $y\in\mathbb{R}^n\setminus\mathcal{L}(x_d,c_k)$ and $x_d^{\pm1}\in\mathcal{PL}(x_d,c_k,y)$. We should also mention that the case where the robot is initially in the set $\Tilde{\mathcal{F}}_0$ is omitted since the robot will operate in the {\it motion-to-destination} mode until it enters the {\it active region} of an obstacle or converges to the destination if the line of sight to the destination is clear for the robot. The generated trajectory is then a line segment. An example illustrating the effect of selecting the virtual destinations as per Lemma \ref{lem6} is shown in Fig. \ref{fig:Lemma5_v}. In Fig.\ref{fig:Lemma5_v}-\subref{fig:fig5_left}, the virtual destinations belong to the two-dimensional plane $\mathcal{PL}(x_d,c_k,x(t_0^k,j_0^k))$, which results in a two-dimensional motion that takes place in the plane $\mathcal{PL}(x_d,c_k,x(t_0^k,j_0^k))$ for both modes considering obstacle $k$ ({\it obstacle-avoidance} mode represented by the orange curve and {\it motion-to-destination} mode represented by the blue curve). In Fig. \ref{fig:Lemma5_v}-\subref{fig:fig5_middle}, the virtual destinations do not belong to the plane $\mathcal{PL}(x_d,c_k,x(t_0^k,j_0^k))$. The motion in the two modes occurs in two different planes, as shown in Fig. \ref{fig:Lemma5_v}-\subref{fig:fig5_right}, where the orange curve represents the {\it obstacle-avoidance} mode, and the blue curve represents the {\it motion-to-destination} mode.\\
When the robot's position belongs to the region $\bigl(\mathcal{J}_k^0\cap\mathcal{C}_k\bigr)$, the jump maps \eqref{jump_map_M}-\eqref{jump_map_B} enable the mode $m$ to take the value $1$ or $-1$ (indistinguishably) when switching from the {\it motion-to-destination} mode to the {\it obstacle-avoidance} mode ({\it i.e.,} avoid the obstacle considering, indistinguishably, the virtual destination $x_k^1$ or $x_k^{-1}$). Leveraging this property together with the fact that robot's motion, during every obstacle avoidance maneuver, is planar (as per Lemma \ref{lem6}), one can 
force the jump from the {\it motion-to-destination} mode, when  $x\in \bigl(\mathcal{J}_k^0\cap\mathcal{C}_k\bigr)$, to the {\it obstacle-avoidance} mode corresponding to the virtual destination closest to the robot's position. This will ensure a smooth transition from the {\it obstacle-avoidance} mode to the {\it motion-to-destination} mode, while guaranteeing locally optimal obstacle avoidance maneuvers as it will be shown later in Proposition \ref{proposition1}.
Figure \ref{figProposition}-\subref{fig_prop1} clearly shows that the green trajectory, generated by switching to the closest virtual destination, $x_k^{-1}$, to the robot's position when first entering the hysteresis region  $\bigl(\mathcal{J}_k^0\cap\mathcal{C}_k\bigr)$ (pink region), is shorter than the blue trajectory generated by selecting $x_k^1$. Figure \ref{figProposition}-\subref{fig_prop2} also shows that when selecting $x_k^{-1}$, the mode switches back to the {\it motion-to-destination} mode earlier than when selecting $x_k^{1}$, as $x_k^{-1}$ becomes visible to the robot before $x_k^{1}$ does. Another observation from Fig. \ref{figProposition}-\subref{fig_prop2} is that when switching back to the {\it motion-to-destination} mode, the robot's position, the destination $x_d$, and the virtual destination $x_k^{-1}$ are aligned, which ensures the continuity of the velocity. As per the discussion above, we propose a modified version of the jump map $B(\cdot)$, defined in \eqref{jump_map_B}, as follows:
\begin{align}\label{jump_map_B_hat}
\small
    \hat{B}(x,k):=\begin{cases}
    1&x\in \mathcal{C}_{k}^{1}\cup\bigl(\mathcal{C}_k\cap\mathcal{P}_{<}(c_k,x_k^{-1}-x_k^1)\bigr),\\
    -1 &x\in \mathcal{C}_{k}^{-1}\cup\bigl(\mathcal{C}_k\cap\mathcal{P}_{>}(c_k,x_k^{-1}-x_k^1)\bigr),\\
    \{-1,1\}&x\in \mathcal{C}_{k}\cap\mathcal{P}_{=}(c_k,x_k^{-1}-x_k^1).
      \end{cases}
\end{align}

Note that by considering the modified jump map $\hat{B}(\cdot)$ in the mode switching scheme, the hysteresis region $\bigl(\mathcal{J}_k^0\cap\mathcal{C}_k\bigr)$, when switching from the {\it motion-to-destination} mode to the {\it obstacle-avoidance} mode, reduces to a line segment (shown in red in Fig. \ref{figProposition}) in the 2D case, thus losing the robustness of the hybrid system. The updating scheme of the mode $m$ in the jump set of the {\it motion-to-destination} mode, considering the modified jump map \eqref{jump_map_B_hat}, and the design of the virtual destinations in Lemma \ref{lem6} are summarized in Algorithm \ref{alg}, and the obtained result is stated in the following proposition.
\begin{proposition}\label{proposition1}
    If the virtual destinations are designed as in Lemma \ref{lem6} and the modified operation mode switching scheme \eqref{jump_map_M}-\eqref{jump_map_B_hat} is considered, the hybrid closed-loop system \eqref{Cl_ctrl} generates continuous velocity control inputs and locally optimal obstacle avoidance maneuvers.
\end{proposition}
\begin{proof}
See Appendix \ref{appendix:proposition1}.
\end{proof}

\begin{algorithm}
 \caption{Mode selector updating scheme in the jump set of the {\it motion-to-destination} mode}\label{alg}
 \begin{algorithmic}[1]
 \renewcommand{\algorithmicrequire}{\textbf{Initialization:}}
 \REQUIRE $x,x_d,\,k,\,c_k,\,r_k,\,e_k,\,\Bar{r}$;
 \ENSURE  $m$.
 \IF{$x\notin\mathcal{L}(x_d,c_k)$}
 \STATE $y\leftarrow x$;
 \ELSE
 \STATE Pick $y\in \mathbb{R}^n\setminus\mathcal{L}(x_d,c_k)$;
 \ENDIF
 \STATE Select the virtual destinations such that\\
 $x_k^1\in\mathcal{H}_k(x_d)\cap\mathcal{P}_{\geq}(p_k,x_d-p_k)\cap\mathcal{PL}(x_d,c_k,y)$\\
 $\|x_d-x_k^1\|=e_k$\\
 $x_k^{-1}\leftarrow x_d-\pi_r(\frac{c_k-x_d}{\|c_k-x_d\|})(x_k^1-x_d)$;
\IF{$x\in\mathcal{C}_k$}
    \IF{$x\in\mathcal{P}_{<}(c_k,x_k^{-1}-x_k^1)$}
    \STATE $m\leftarrow 1$;
    \ELSIF{$x\in\mathcal{P}_{>}(c_k,x_k^{-1}-x_k^1)$}
    \STATE $m\leftarrow -1$;
    \ELSE
    \STATE $m\leftarrow \{-1,1\}$;
    \ENDIF
\ELSE
\STATE $m\leftarrow B(x,k)$ using \eqref{jump_map_B};
\ENDIF
\RETURN m;
 \end{algorithmic} 
 \end{algorithm}
The implementation of the hybrid control \eqref{hyb_ctrl} is summarized in Algorithm \ref{alg1}, where the steps colored in blue are only required for the sensor-based implementation discussed in the following section. For compactness, we write the workspace's data and design parameters as follows: \(c:=[c_1,\dots,c_b]\), \(r:=[r_1,\dots,r_b]^{\top}\), \(\Bar{r}:=[\Bar{r}_1,\dots,\Bar{r}_b]^{\top}\) and \( e:=[e_1,\dots,e_b]^{\top}\).

\begin{algorithm}
 \caption{General implementation of the hybrid control \eqref{hyb_ctrl}}\label{alg1}
 \begin{algorithmic}[1]
 \renewcommand{\algorithmicrequire}{\textbf{Initialization:}}
 \REQUIRE $x_d,\,e_c,\,c,\,r,\,\Bar{r},\,e,\,\epsilon$, {\color{blue} $e_s$}, $x(0,0)\in\mathcal{X}$, $k(0,0)\in\mathbb{I}$, and $m(0,0)\in\mathbb{M}$;
 \WHILE{true}
 \STATE Measure $x$;
 \IF{$\|x-x_d\|\leq e_c$}
 \STATE Break;
 \ELSE
 {\color{blue}
 \STATE Obtain the list of arcs $\mathcal{R}(x)$;
 \STATE Reconstruct obstacles using \eqref{reconst_proj}-\eqref{reconst_center_2D};
 \STATE Dilate obstacles radii: $r_i\leftarrow r_i+e_s,\,i\in\mathbb{S}_x$;
 \STATE Update $\Bar{r}$;
 }
 \IF{$(x,m)\in\mathring{\mathcal{J}}_0\times\{0\}$}
 \STATE Update $k$ using \eqref{jump_map_K}; 
 \STATE Update $m$ using Algorithm \ref{alg};
 \ELSE
\STATE Update $k$ and $m$ using \eqref{jump_dyn_c}-\eqref{jump_dyn};
\ENDIF
\STATE Execute $u(x,k,m)$ using \eqref{hyb_ctrl_L};
\ENDIF
\ENDWHILE
 \end{algorithmic} 
 \end{algorithm}
\section{Sensor-Based Implementation}\label{section V}
Since the workspace is assumed to contain spherical obstacles, one can reconstruct the obstacles from their detected portions obtained via a range scanner that covers a region $\mathcal{B}(x,R),\, R>0,$ around the robot. As the detection region is limited to the sensor's range $R$, we redefine the range of the {\it active region} of obstacle $k\in\mathbb{I}$, defined in \eqref{active_region}, by 
\begin{align}\label{redefine_active_range}
    \Bar{r}_k\in(0,\Tilde{r}_k),\;\Tilde{r}_k:=\min (\hat{r}_k,R).
\end{align}
Next, we implement our hybrid strategy in two and three-dimensional spaces using 2D and 3D LiDAR range scanners ({\it e.g.,} LEICA, BLK, ARC  scanning modules).  
\subsection{Two-dimensional spaces}
\begin{figure}[h!]
\centering
\includegraphics[scale=0.4]{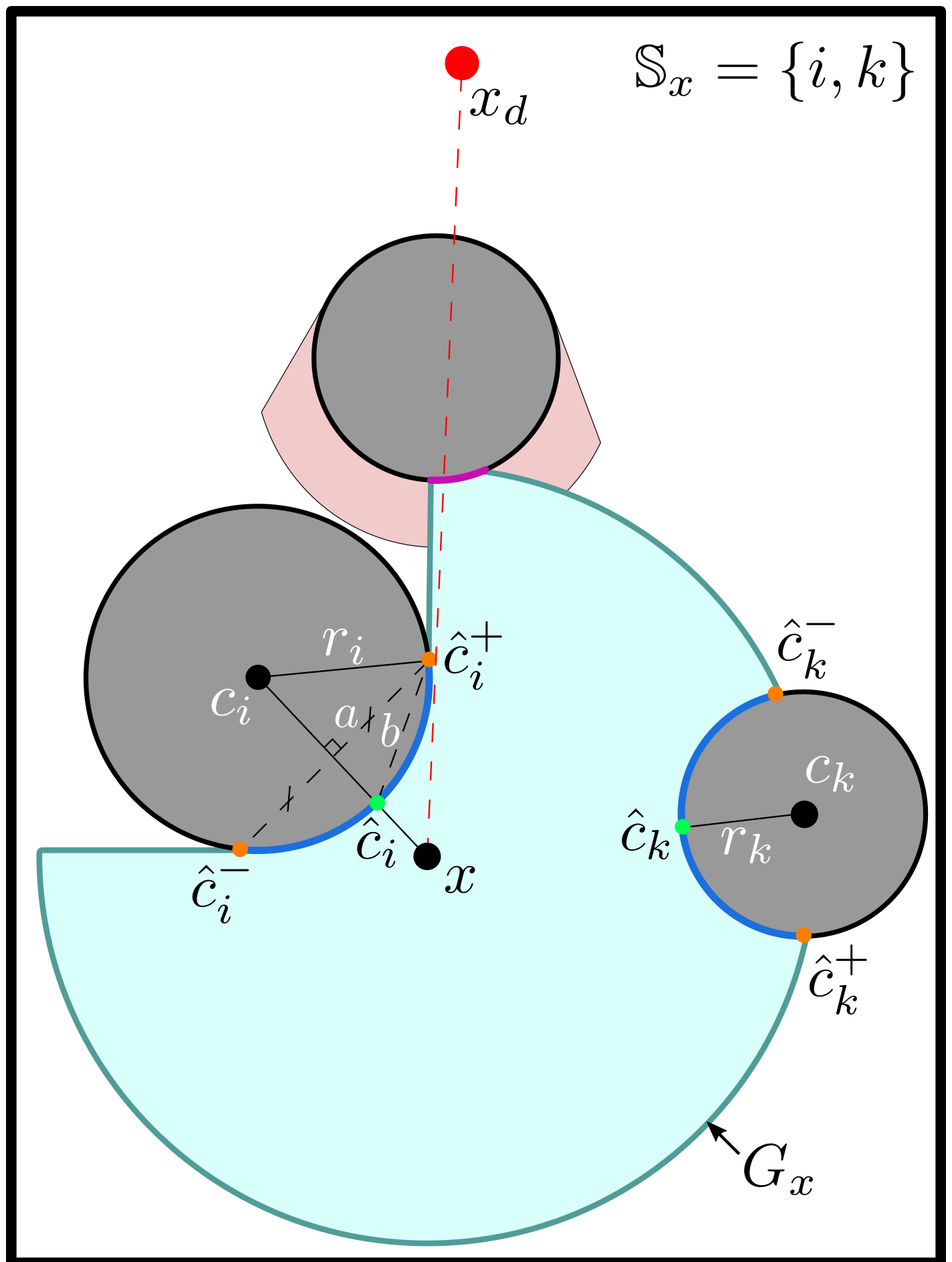}
\caption{Obstacle reconstruction from sensor data.}
\label{fig6}
\end{figure}

 Consider a two-dimensional workspace, and assume that the robot is equipped with a LiDAR of resolution $d\psi>0$, a maximum radial range $R>0$, and an angular range of $360^{\circ}$. We model the measurements of the sensor, at a position $x$, by the polar curve $\rho(x,\psi):\mathcal{X}\times\mathcal{G}\rightarrow[0, R]$ defined as follows:
 {\small
\begin{align}
    \rho(x,\psi):=\min\left(\begin{array}{l}
        R,
        \min\limits_{\substack{y\in\partial\mathcal{X}\\\mathrm{atan2}(y(2)-x(2),y(1)-x(1))=\psi}}\|y-x\|
    \end{array}\right),
\end{align}
}
where $\mathcal{G}:=\left\{0,d\psi,2d\psi,\dots,360-d\psi\right\}$ is the set of scanned angles. The Cartesian coordinates of the scanned points are modeled by the mapping  
 $\delta(x,\psi):\mathcal{X}\times\mathcal{G}\rightarrow\mathcal{X}$ defined as follows:
 \begin{align}
    \delta(x,\psi):=x+\rho(x,\psi)[\cos(\psi)\;\sin(\psi)]^{\top}.
\end{align}
Let $G_x(\delta)$ be the graph of the mapping $\delta$ at a position $x$. The set $\mathbb{I}_x\subset\mathbb{I}$ of the detected obstacles, at position $x$, is defined as $\mathbb{I}_x:=\left\{i\in\mathbb{I}|d(x,\mathcal{O}_i)\leq R\right\}$. Assume that at each position $x$, the sensor returns a list of arcs $\mathcal{R}(x):=\left\{A_1,A_2,\dots,A_{\iota(x)}\right\}$ from the detected obstacles corresponding to the intersection of the graph $G_x(\delta)$ and obstacles of the set $\mathbb{I}_x$, as shown in Fig. \ref{fig6}, where $\iota(x)=\textbf{card}(\mathbb{I}_x)$. Using the arcs of the list $\mathcal{R}(x)$, at a position $x$, one can reconstruct the obstacles by determining their centers and radii. Due to the LiDAR's radial sweep, at positions where some obstacles are partially hidden by other obstacles, the detected arcs of partially hidden obstacles may be asymmetrical with respect to the projection of the robot's position on these obstacles. These asymmetrical arcs are ignored as they imply that the robot is outside the {\it active region} of their associated obstacles since, according to definitions \eqref{active_region} and \eqref{redefine_active_range}, the {\it active region} must be free of any other obstacles (see also Fig. \ref{fig6}). The indices of obstacles associated with the symmetrical arcs, detected at position $x$, are grouped in the set $\mathbb{S}_x\subset\mathbb{I}_x$. Consider a symmetrical arc $A_i\in\mathcal{R}(x)$ associated with obstacle $\mathcal{O}_k$ where $i\in\{1,\dots,\iota(x)\}$ and $k\in\mathbb{S}_x$. The center $c_k$ and the radius $r_k$ can be obtained, as illustrated in Fig. \ref{fig6}, through the following steps:
\begin{itemize}
    \item Determine the projection of $x$ onto the arc $A_i$ ({\it i.e.,} the closest point of obstacle $k$ to the robot) as follows:
    \begin{align}\label{reconst_proj}
        \hat{c}_i:=\arg \min\limits_{y\in A_i} \|x-y\|.
    \end{align}
    \item Determine the radius $r_k$ as follows:
    \begin{align}\label{reconst_radius_2D}
        r_k:=\frac{b^2}{2\sqrt{b^2-a^2}},
    \end{align}
    where $a=\|c_i^{+}-c_i^{-}\|$, $b=\|c_i^{+}-\hat{c}_i\|=\|c_i^{-}-\hat{c}_i\|$, and $c_i^{+}$, $c_i^{-}$ are the endpoints of arc $A_i$.
    \item Determine the center $c_k$ as follows:
    \begin{align}\label{reconst_center_2D}
        c_k:=\hat{c}_i+r_k\frac{\hat{c}_i-x}{\|\hat{c}_i-x\|}.
    \end{align}
\end{itemize}
Since the centers and radii of the detected obstacles in the vicinity of the robot can be determined, the hybrid control \eqref{hyb_ctrl} can be implemented in an unknown two-dimensional workspace with disc-shaped obstacles, as described in Algorithm \ref{alg1} considering the steps colored in blue.
\subsection{Three-dimensional spaces}
Consider a three-dimensional workspace, and assume that the robot is equipped with a 3D-LiDAR of polar resolution $d\vartheta>0$, polar angular range of $180^{\circ}$, azimuthal resolution $d\psi>0$, azimuthal angular range of $360^{\circ}$, and a maximum radial range $R>0$. We model the measurements of the sensor, at a position $x$, by the curve $\Bar{\rho}(x,\vartheta,\psi):\mathcal{X}\times\mathcal{U}\times\mathcal{G}\rightarrow[0, R]$ defined as follows:
{\small
\begin{align}
    \Bar{\rho}(x,\vartheta,\psi):=\min\left(\begin{array}{l}
        R,
        \min\limits_{\substack{y\in\partial\mathcal{X}\\\mathrm{atan2}(y(2)-x(2),y(1)-x(1))=\psi\\
        \arccos(\|y-x\|/(y(3)-x(3)))=\vartheta}}\|y-x\|
    \end{array}\right),
\end{align}
}
where $\mathcal{U}:=\{0,d\vartheta,2d\vartheta,\dots,360-d\vartheta\}$ and $\mathcal{G}:=\left\{0,d\psi,2d\psi,\dots,180-d\psi\right\}$ are, respectively, the set of scanned polar angles and the set of scanned azimuthal angles. The Cartesian coordinates of the scanned points are modeled by the mapping  
 $\Bar{\delta}(x,\vartheta,\psi):\mathcal{X}\times\mathcal{G}\times\mathcal{U}\rightarrow\mathcal{X}$ defined as follows:
 \begin{multline}
    \Bar{\delta}(x,\vartheta,\psi):=x\\+\Bar{\rho}(x,\vartheta,\psi)[\cos(\psi)\sin(\vartheta)\;\sin(\psi)\sin(\vartheta)\;\cos(\vartheta)]^{\top}.
\end{multline}
Similar to the two-dimensional case, $G_x(\Bar{\delta})$ represents the graph of the mapping $\Bar{\delta}$ at a position $x$ and $\mathbb{I}_x\subset\mathbb{I}$ the set of detected obstacles at position $x$. Assume that at each position $x$, the sensor returns a list of spherical caps $\Bar{\mathcal{R}}(x):=\left\{\Bar{A}_1,\Bar{A}_2,\dots,\Bar{A}_{\iota(x)}\right\}$ from the detected obstacles corresponding to the intersection of the graph $G_x(\Bar{\delta})$ and obstacles of the set $\mathbb{I}_x$, where $\iota(x)=\textbf{card}(\mathbb{I}_x)$. The intersection of two spheres is a circle if the distance between their centers is less than the sum of their radii and if neither sphere is enclosed by the other. As a result, the detected spherical caps are formed by a bump with a circular base. At certain positions, some obstacles may be partially hidden from the sensor by other obstacles. The detected parts of the partially hidden obstacles may be asymmetrical with respect to the projection of the robot's position onto these obstacles. At positions where asymmetrical spherical caps are detected, the robot is certainly outside the {\it active regions}  of obstacles associated with the asymmetrical caps since, as per definition \eqref{active_region}-\eqref{redefine_active_range}, the {\it active region} of an obstacle should not contain any other obstacle. Consequently, we ignore the asymmetrical spherical caps as their associated obstacles are not required for the control. The indices of obstacles associated with the symmetrical spherical caps, detected at position $x$, are grouped in the set $\mathbb{S}_x\subset\mathbb{I}_x$. Consider a symmetrical spherical cap $\Bar{A}_i\in\Bar{\mathcal{R}}(x)$, detected at position $x\in\mathcal{X}$, associated with obstacle $\mathcal{O}_k$ where $i\in\{1,\dots,\iota(x)\}$ and $k\in\mathbb{S}_x$. The reconstruction of obstacle $\mathcal{O}_k$ from its detected spherical cap $\Bar{A}_i$ can be obtained as follows:
\begin{itemize}
    \item Determine the projection of $x$ onto the spherical cap $\Bar{A}_i$ ({\it i.e.}, the closest point of obstacle $\mathcal{O}_k$ to the robot)
    \begin{align}
        \Tilde{c}_i:=\arg \min\limits_{y\in \Bar{A}_i} \|x-y\|.
    \end{align}
    \item Determine the radius of obstacle $\mathcal{O}_k$
    \begin{align}
        r_k:=\frac{b^2}{2\sqrt{b^2-a^2}},
    \end{align}
    where $a$ is the radius of the circular basis $\Bar{\mathcal{C}}_i$ of the portion $\Bar{A}_i$, and $b=d(\Tilde{c}_i,\Bar{\mathcal{C}}_i)$.
    \item Determine the center of obstacle $\mathcal{O}_k$
    \begin{align}
        c_k:=\Tilde{c}_i+r_k\frac{\Tilde{c}_i-x}{\|\Tilde{c}_i-x\|}.
    \end{align}
\end{itemize}
Now, with the information of neighboring obstacles to the robot available through the sensor's output, the hybrid control \eqref{hyb_ctrl} can be implemented in unknown spherical three-dimensional spaces as summarized in Algorithm \ref{alg1}, considering the steps colored in blue and replacing $\mathcal{R}(x)$ in step 6 of the algorithm with $\Bar{\mathcal{R}}(x)$.
\begin{rem}
     For safer navigation, the numerical errors and low resolution of the LiDAR should be considered. Therefore, a security margin $e_s>0$ can be added to the radius of the detected obstacles where the separation between every pair of obstacles has to be larger than $2 e_s$ ({\it i.e.,} $\forall i,k\in\mathbb{I},\,i\neq k,\,\|c_k-c_i\|-r_k-r_i>2e_s$.) 
 \end{rem}
\section{Numerical Simulation}  
\subsection{Implementation with global knowledge of the environment}\label{gen_sim}
\begin{figure}[h!]
\centering
\includegraphics[width=\columnwidth]{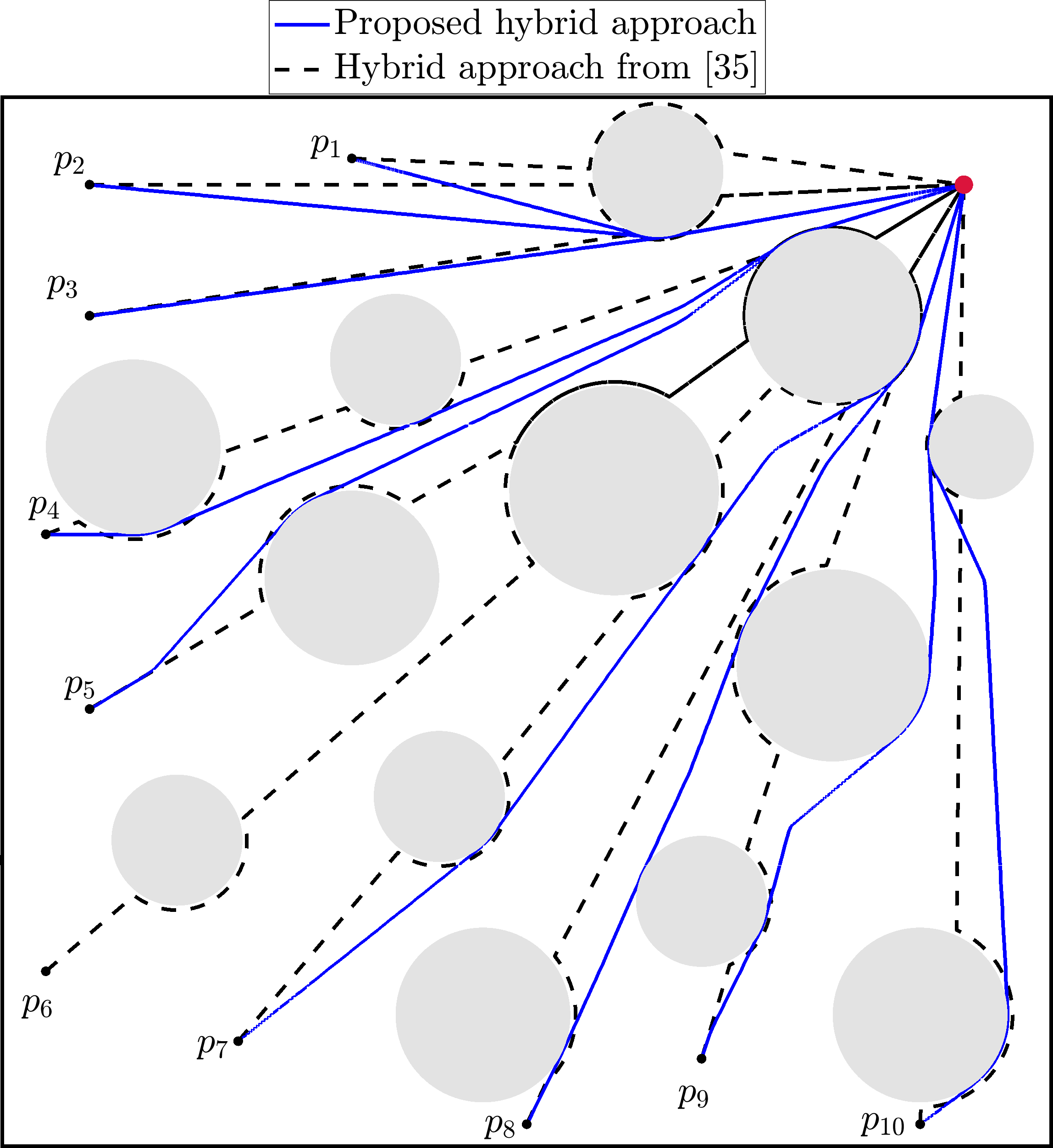}
\caption{Robot navigation trajectories from ten different initial positions in a 2D workspace cluttered with circular obstacles. The blue trajectories are generated by the proposed hybrid approach and the black trajectories are generated by the hybrid approach from \cite{SoulaimaneHybTr}. The target is represented by the red dot.}
\label{sim1}
\end{figure}

\begin{figure}[h!]
\centering
\includegraphics[width=\columnwidth]{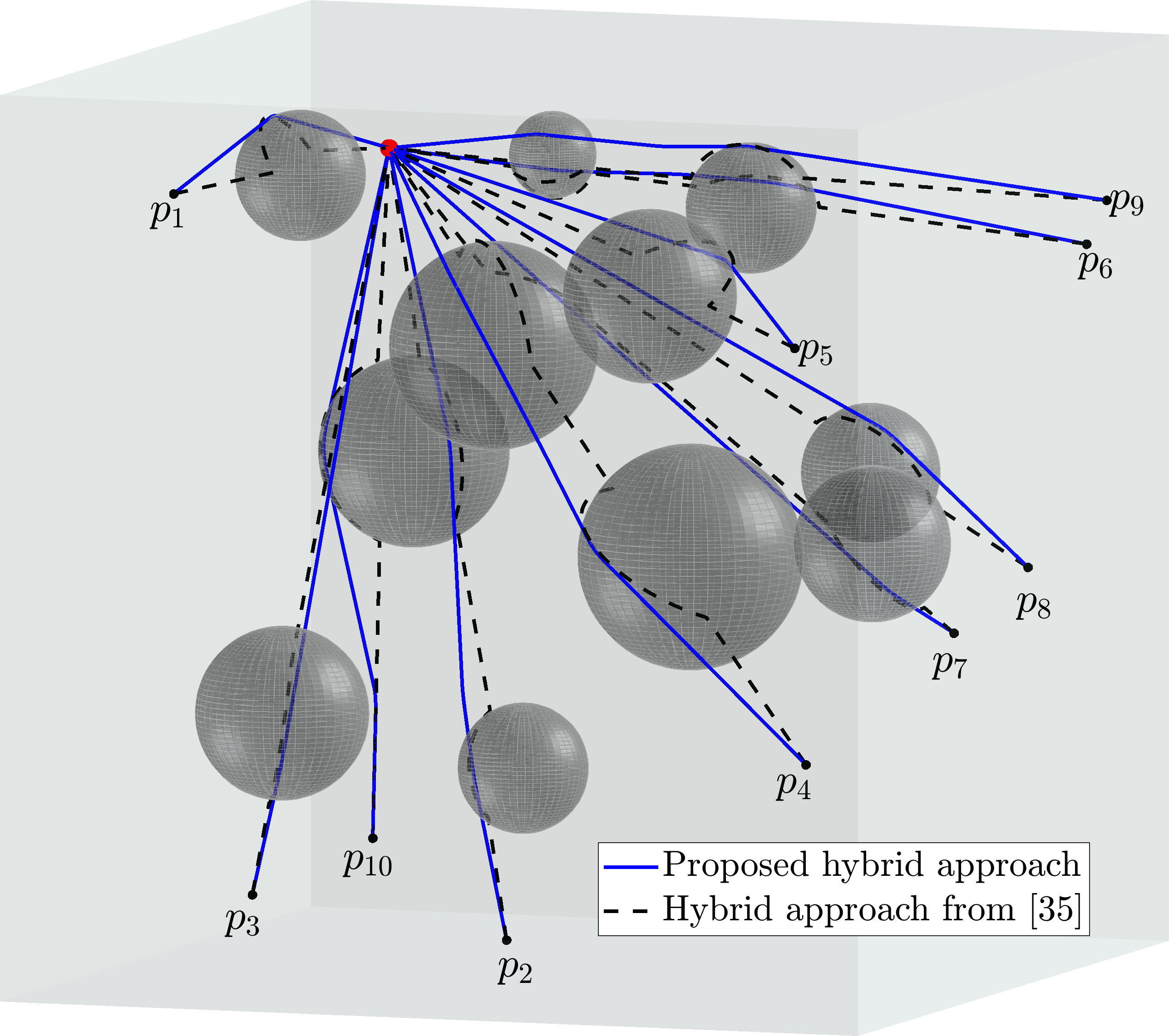}
\caption{Robot navigation trajectories from ten different initial positions in a 3D workspace cluttered with spherical obstacles. The blue trajectories are generated by the proposed hybrid approach and the black trajectories are generated by the hybrid approach from \cite{SoulaimaneHybTr}. The target is represented by the red dot.}
\label{sim2}
\end{figure}

In order to visualize the performance of our proposed hybrid approach, we compare it with another hybrid approach that considers a single integrator model and guarantees safety and GAS in $n$-dimensional Euclidean spaces, proposed in \cite{SoulaimaneHybTr}. We performed simulations starting from 10 different initial conditions in two different workspaces. The first experiment is done in a two-dimensional environment, as shown in Fig. \ref{sim1}, where we plotted the trajectories obtained by our approach along with the trajectories generated by the approach proposed in \cite{SoulaimaneHybTr}. We also report the relative length difference of the paths generated by the approach proposed in \cite{SoulaimaneHybTr} with respect to ours in Table \ref{table_2D}. For each initial position $p_i$, $i\in\{1,\dots,10\}$, in Fig. \ref{sim1}, we computed the relative length difference $RLD_i=100(L_i-l_i)/l_i$, where $L_i$ is the length of the $i$th path generated by the approach proposed in \cite{SoulaimaneHybTr}, and $l_i$ is the length of the path generated by our approach. The trajectories plotted in Fig. \ref{sim1} show clearly that our approach generates a continuous control input (robot's velocity) while the approach proposed in \cite{SoulaimaneHybTr} generates a discontinuous control input. We can also observe that our trajectories are shorter than the ones generated by the approach proposed in \cite{SoulaimaneHybTr}, which is confirmed by the positive relative difference reported in Table \ref{table_2D}. This difference in length is mainly due to the fact that our approach starts the {\it obstacle-avoidance} mode, with local optimal maneuvers, once in the {\it active region} of an obstacle and switches back to the {\it motion-to-destination} once the avoided obstacle is no longer blocking the view of the destination $x_d$. In contrast, the approach proposed in \cite{SoulaimaneHybTr} starts the {\it obstacle-avoidance} mode in close vicinity of the obstacle, performing a boundary-following motion on a helmet covering the obstacle. It does not switch back to the {\it motion-to-destination} mode once the avoided obstacle stops blocking the view of the destination $x_d$, but once it exits the helmet. To ensure that the performance of our approach is preserved regardless of the dimension of the workspace, we repeated the same experiment in a three-dimensional environment and reported the results in Fig. \ref{sim2} and Table \ref{table_3D}. The same observations can be drawn from this experiment as in the 2D case, concluding the efficiency of our approach in higher dimensions. Notably, the approach proposed in \cite{SoulaimaneHybTr} considers a more general class of obstacles, namely ellipsoids, whereas our approach only considers spheres. Nevertheless, our approach can be implemented in {\it a priori} unknown environments using only on-board range scanners as illustrated in the following simulations.
\begin{table}[h!]
\caption{The relative length difference of the paths, shown in Fig. \ref{sim1}, generated by the hybrid approach from \cite{SoulaimaneHybTr} with respect to the proposed hybrid approach in a 2D workspace.
}
\label{table_2D}
\begin{center}
\resizebox{1\linewidth}{!}{
\begin{tabular}{|c||c||c||c||c||c||c||c||c||c||c|}
\hline
\textbf{Paths} & $\boldsymbol{p_1}$& $\boldsymbol{p_2}$& $\boldsymbol{p_3}$& $\boldsymbol{p_4}$& $\boldsymbol{p_5}$& $\boldsymbol{p_6}$& $\boldsymbol{p_7}$& $\boldsymbol{p_8}$& $\boldsymbol{p_9}$& $\boldsymbol{p_{10}}$\\
\hline
$RLD$ (\%) &7.36& 4.39& 1.76&7.33& 10.68& 9.97& 11.2& 3.95& 6.08& 5.14 \\
\hline
\end{tabular}}
\end{center}
\end{table}
\begin{table}[h!]
\caption{The relative length difference of the paths, shown in Fig. \ref{sim2}, generated by the hybrid approach from \cite{SoulaimaneHybTr} with respect to the proposed hybrid approach in a 3D workspace.
}
\label{table_3D}
\begin{center}
\resizebox{1\linewidth}{!}{
\begin{tabular}{|c||c||c||c||c||c||c||c||c||c||c|}
\hline
\textbf{Paths} & $\boldsymbol{p_1}$& $\boldsymbol{p_2}$& $\boldsymbol{p_3}$& $\boldsymbol{p_4}$& $\boldsymbol{p_5}$& $\boldsymbol{p_6}$& $\boldsymbol{p_7}$& $\boldsymbol{p_8}$& $\boldsymbol{p_9}$& $\boldsymbol{p_{10}}$\\
\hline
$RLD$ (\%) &6.38& 4.91& 4.18&8.78& 5.02& 5.71& 5.12& 4.9& 7.15& 5.25 \\
\hline
\end{tabular}}
\end{center}
\end{table}
\subsection{Sensor-based implementation}
\begin{figure}[h!]
\centering
\includegraphics[width=\columnwidth]{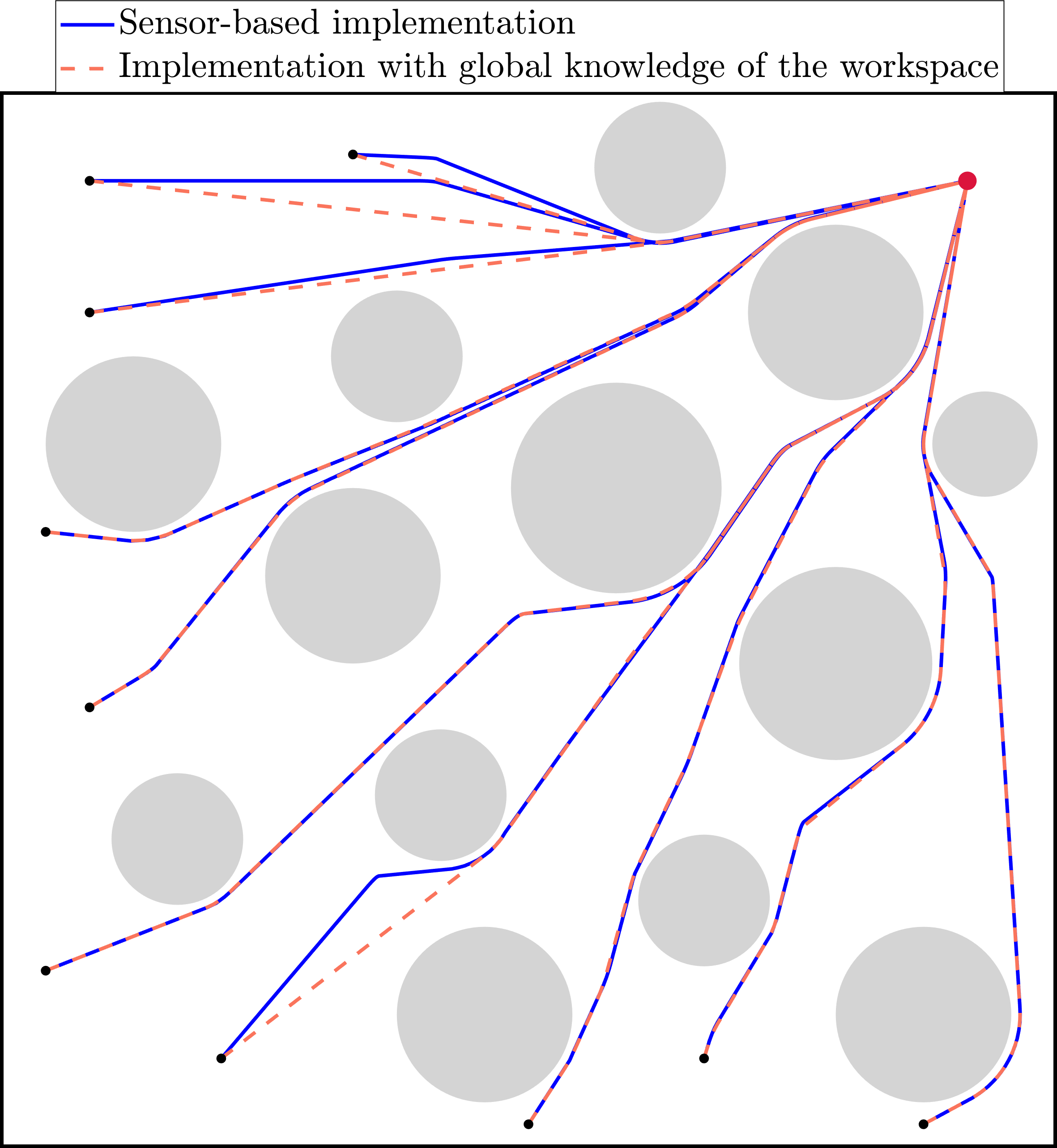}
\caption{Robot navigation trajectories from ten different initial positions in a 2D workspace cluttered with circular obstacles. The blue trajectories are generated by the sensor-based implementation of the proposed approach, and the orange trajectories are generated when the environment is {\it a priori} known. The red dot represents the target.}
\label{sim3}
\end{figure}

\begin{figure}[h!]
\centering
\includegraphics[width=\columnwidth]{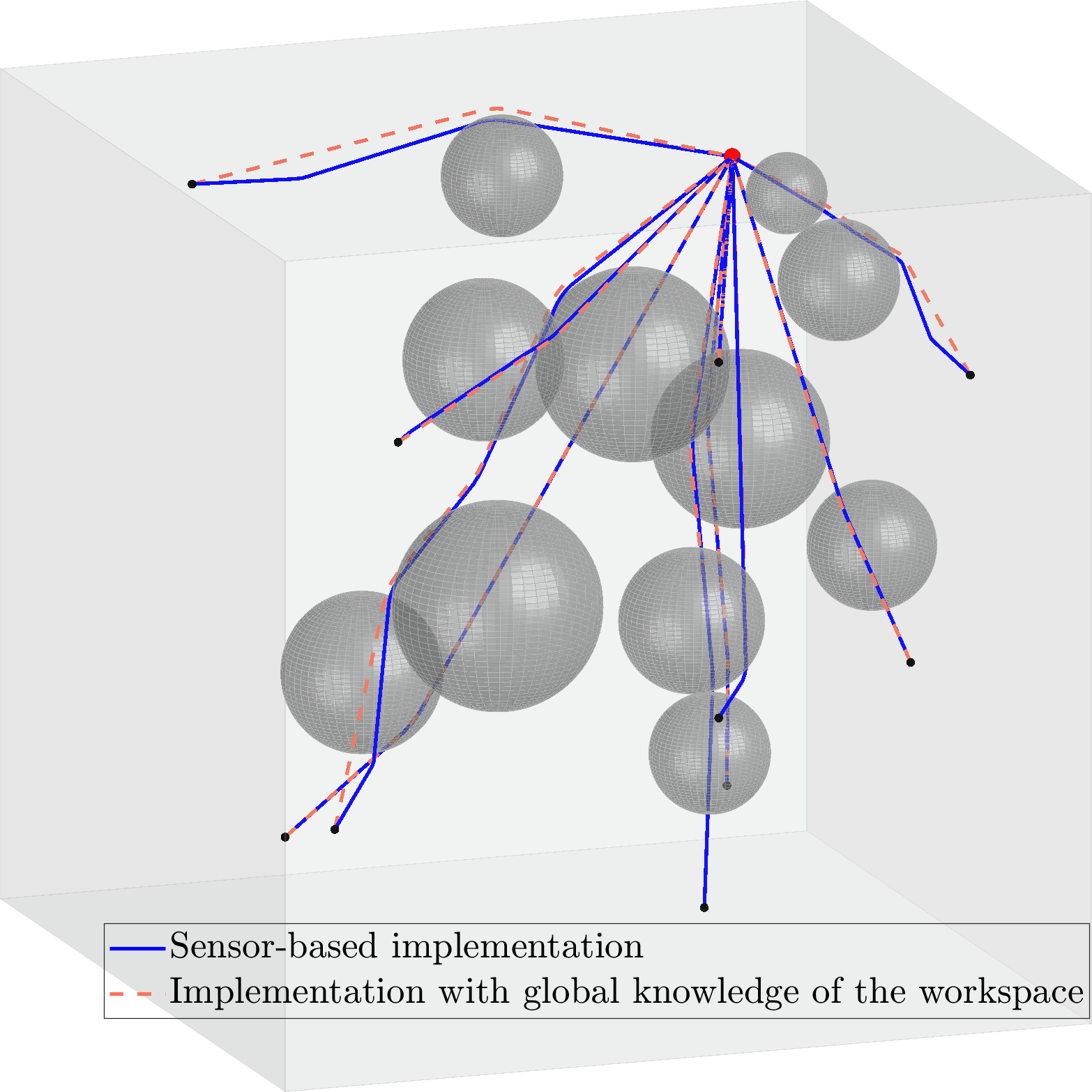}
\caption{Robot navigation trajectories from ten different initial positions in a 3D workspace cluttered with spherical obstacles. The blue trajectories are generated by the sensor-based implementation of the proposed approach, and the orange trajectories are generated by when the environment is {\it a priori} known. The red dot represents the target.}
\label{sim4}
\end{figure}

To test the practicality of our approach, we simulated the sensor-based implementation in the same spaces as in the general implementation section \ref{gen_sim}. For the 2D case, we used a $360^{\circ}$-LiDAR model with $0.5^{\circ}$ resolution and $2m$ radial range. For the 3D case, we used a 3D-LiDAR with $1^{\circ}$ polar and azimuthal resolutions, $180^{\circ}$ polar angular range, $360^{\circ}$ azimuthal angular range, and $2m$ radial range. We considered a security margin $e_s=0.1m$ for the obstacles radii. We plotted the trajectories obtained through the sensor-based implementation along with the ones generated when the environment is {\it a priori known} in Fig. \ref{sim3} (resp. Fig. \ref{sim4}) for the 2D case (resp. for the 3D case). As the sensor's range is limited, the {\it active regions} have been redefined in \eqref{redefine_active_range}, which explains the fact that the sensor-based implementation generates paths that are, in some cases, longer than the general implementation. Nevertheless, the sensor-based implementation generally reproduces the same trajectories as the implementation with a global knowledge of the environment. Simulation videos for the 2D case and 3D case showing the sensor-based navigation can be viewed from the links in the footnote\footnote{\url{https://youtube.com/shorts/bL0dOl7W9Ms?feature=share} and \url{https://youtu.be/oJqpUW8Blb4}}.
\section{Experimental validation}\label{experimental_validation}
In this section, we use the Turtlebot 4 platform shown in Fig. \ref{fig:TBT4}(left) to implement the control strategy \eqref{hyb_ctrl} using the following transformation 
\begin{align}\displaystyle
\label{diff_drive_ctrl}
    \begin{cases}
      v=\min\left(v_{\max},k_v\|u(x)\|\left(\cos(\frac{\Delta\Phi}{2})\right)^{2p}\right), \\
      \omega=\omega_{\max}\sin(\frac{\Delta\Phi}{2}),
    \end{cases}
\end{align}
which adapts the control law $u(x)$ used for the fully-actuated model to the control inputs $(v,\omega)$ of the following differential-drive model
\begin{align}\label{differential_drive}
    \begin{cases}
      \Dot{x}=v[\cos(\Phi)\,\sin(\Phi)]^\top, \\
      \Dot{\Phi}=\omega,
    \end{cases}
\end{align}
where $\Phi\in (-\pi,\pi]$ is the robot's orientation, $\Delta \Phi\in(-\pi,\pi]$ is the difference between the robot's orientation and the direction of $u(x)$, $v\in\mathbb{R}$ and $\omega\in\mathbb{R}$ are, respectively, the robot's linear and angular velocity inputs, $v_{\max}=0.31\,m/s$ and $\omega_{\max}=1.9\,rd/s$ are the maximum supported velocities by the robot's actuators, $k_v>0$, and $p\geq1$. Larger $p$ values lead to small velocities when the robot's heading is misaligned with the direction of the control $u(x)$ ({\it i.e., $\Delta\Phi\neq0$}). This minimizes the linear displacements when the robot orients its heading to match $u(x)$. This procedure allows to generate trajectories closer to the ones generated by the control $u(x)$. Since the Turtlebot 4 has a disc-shaped base of radius $r_t = 0.17m$, we dilate the obstacles by a dilation factor $r_d = r_t + r_s$ where $r_s = 0.13m$ is a safety margin.
\subsection{Experimental settings}\label{exp_settings}
\begin{figure}[!h]
\centering
\includegraphics[width=\linewidth]{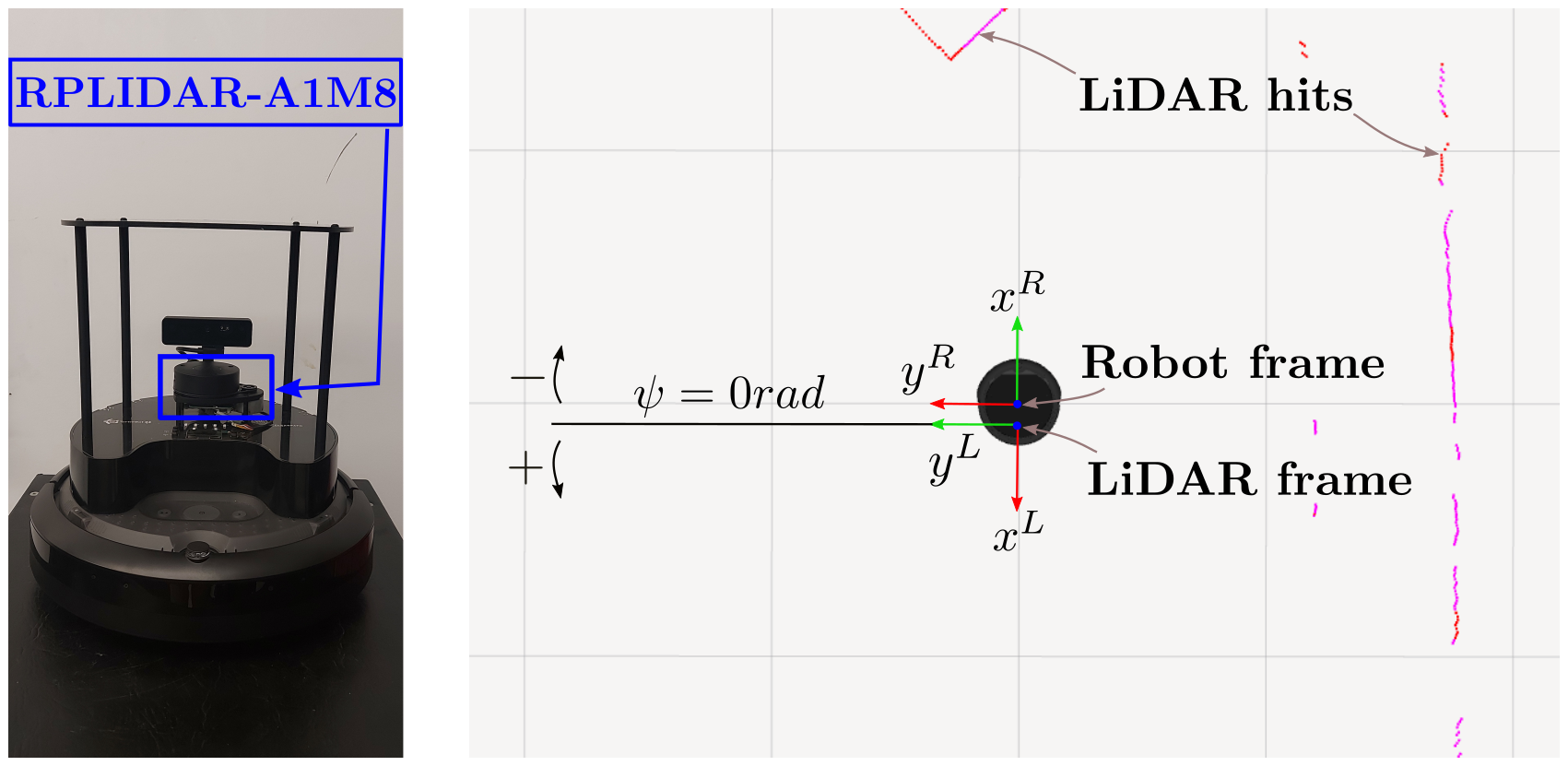}
\caption{The figure on the left shows the Turtlebot 4 and the RPLIDAR-A1M8. The figure on the right illustrates the robot frame and the LiDAR frame.}
\label{fig:TBT4}
\end{figure}

\begin{figure}[!h]
\centering
\includegraphics[width=1\linewidth]{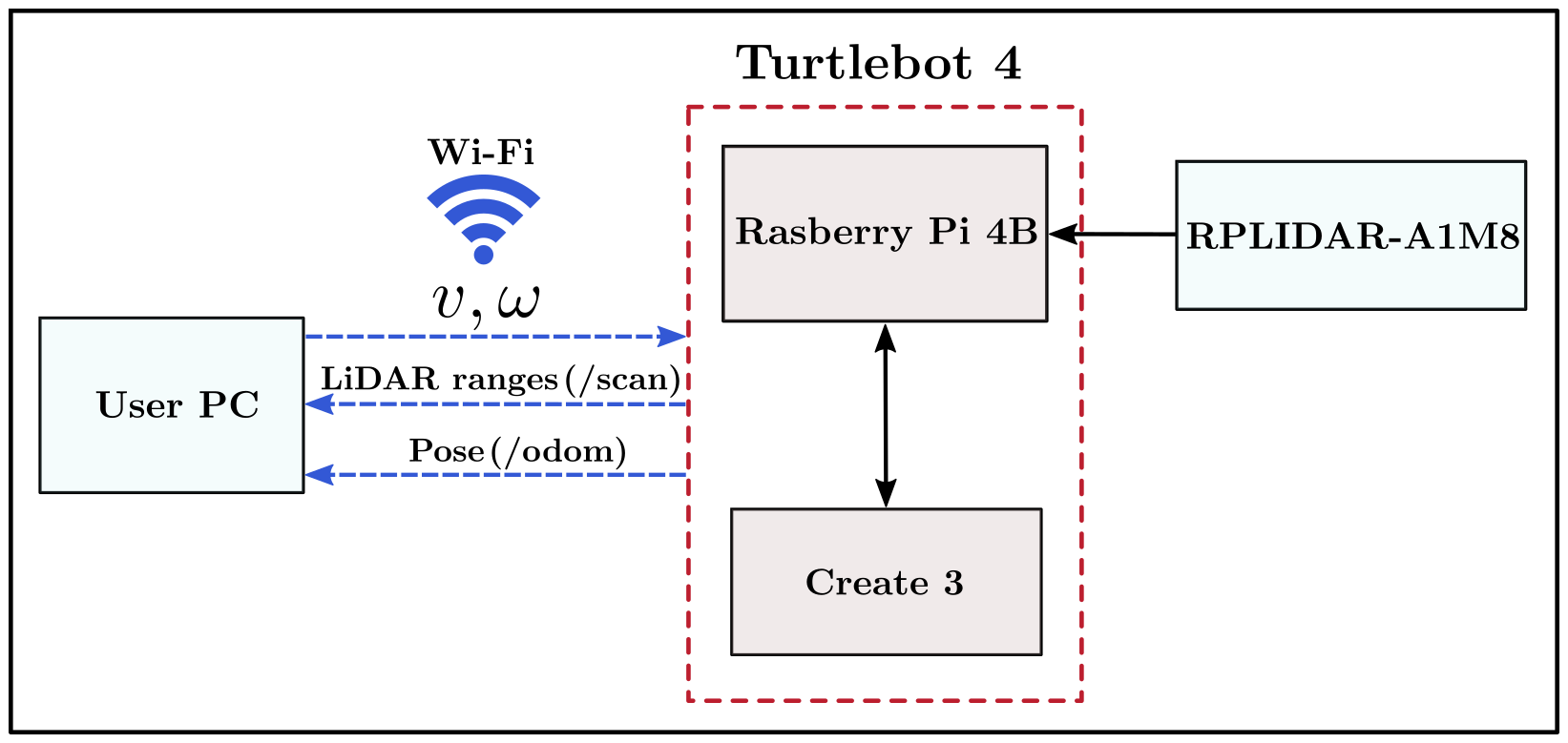}
\caption{A schematic representation of the network and data communication flow in our experimental setup.  
}
\label{fig:Networking}
\end{figure}

Our implementation is based on a ROS 2 (Humble) setup on Ubuntu 22.04, integrating the two main components of the TurtleBot 4, the Create 3 and the Raspberry Pi 4B (RPi4B), and a User PC (an external Intel(R) Core(TM) i5-6500 CPU @ 3.20GHz machine with 8GB RAM), with communication realized through a simple discovery mode. This networking mode is a multicast configuration enabling peer-to-peer communication between the various devices connected to the Wi-Fi network. The RPi4B acts as the main host of the robot's ROS 2 node and as a network gateway. It receives LiDAR data and transmits it to the network, publishing it in the /scan topic. It also relays the robot pose information supplied by Create 3 to the network, publishing it in the /odom topic, and receives the velocity commands sent by the PC user and transmits it to Create 3. The Create 3 is the mobile base equipped with essential onboard actuators and sensors such as wheel encoders and IMU for odometry and cliff sensors for safety. It is also responsible for the low-level control. The user PC receives LiDAR ranges and robot pose to execute the control algorithm and send velocity commands (linear velocity $v$ and angular velocity $\omega$) to the RPi4B, publishing them in the /vel\textunderscore cmd topic. The network and data communication flow are shown in \ref{fig:Networking}. For obstacle detection, we rely on the onboard RPLiDAR-A1MS shown in Fig. \ref{fig:TBT4}. The Lidar has a resolution of $1\deg$, an angular range of $360\deg$, a minimum radial range $R_{min} = 0.15 m$ and a maximum radial range $R_{\max} = 12 m$. The LiDAR provides measurements in its own frame (LiDAR frame), which is displaced by $-4cm$ along the $x$-axis and rotated by $90\deg$ with respect to the robot frame as illustrated in Fig. \ref{fig:TBT4}(right). Therefore, we first limit the measurement radial range to $R_{\max}=1.5m$ (due to the workspace limitation), then transform the LiDAR data from the LiDAR frame to the robot frame so we can correctly localize the detected obstacles with respect to the robot frame. The parameters of the control \eqref{hyb_ctrl} and \eqref{diff_drive_ctrl} used in the experiment are summarized in Table \ref{table_experimental_parameters}.
\begin{table}[h!]
\caption{Experimental setup and control parameters.}
\label{table_experimental_parameters}
\begin{center}
\begin{tabular}{|c||c|}
\hline
Parameter & Value\\
\hline
Robot's radius & $r_t = 0.17m$ \\
\hline
Safety margin & $r_s=0.13m$ \\
\hline
Dilation parameter & $r=r_t+r_s=0.3m$\\
\hline
Maximum radial range measurement & $R_{max}=1.5m$\\
\hline
Gain of the control $u(x)$ & $\gamma = 1.5$ \\
\hline
Gain used in \eqref{hyb_ctrl_L} & $k_v=0.1$\\
\hline
Tuning parameter used in \eqref{diff_drive_ctrl}  & $p=1$  \\
\hline
Maximum linear velocity & $v_{\max}=0.31m/s$ \\
\hline
Maximum angular velocity & $\omega_{\max}=1.9rd/s$\\
\hline
\end{tabular}
\end{center}
\end{table}
\subsection{Experimental results}
\begin{figure}[!h]
\centering
\includegraphics[width=\linewidth]{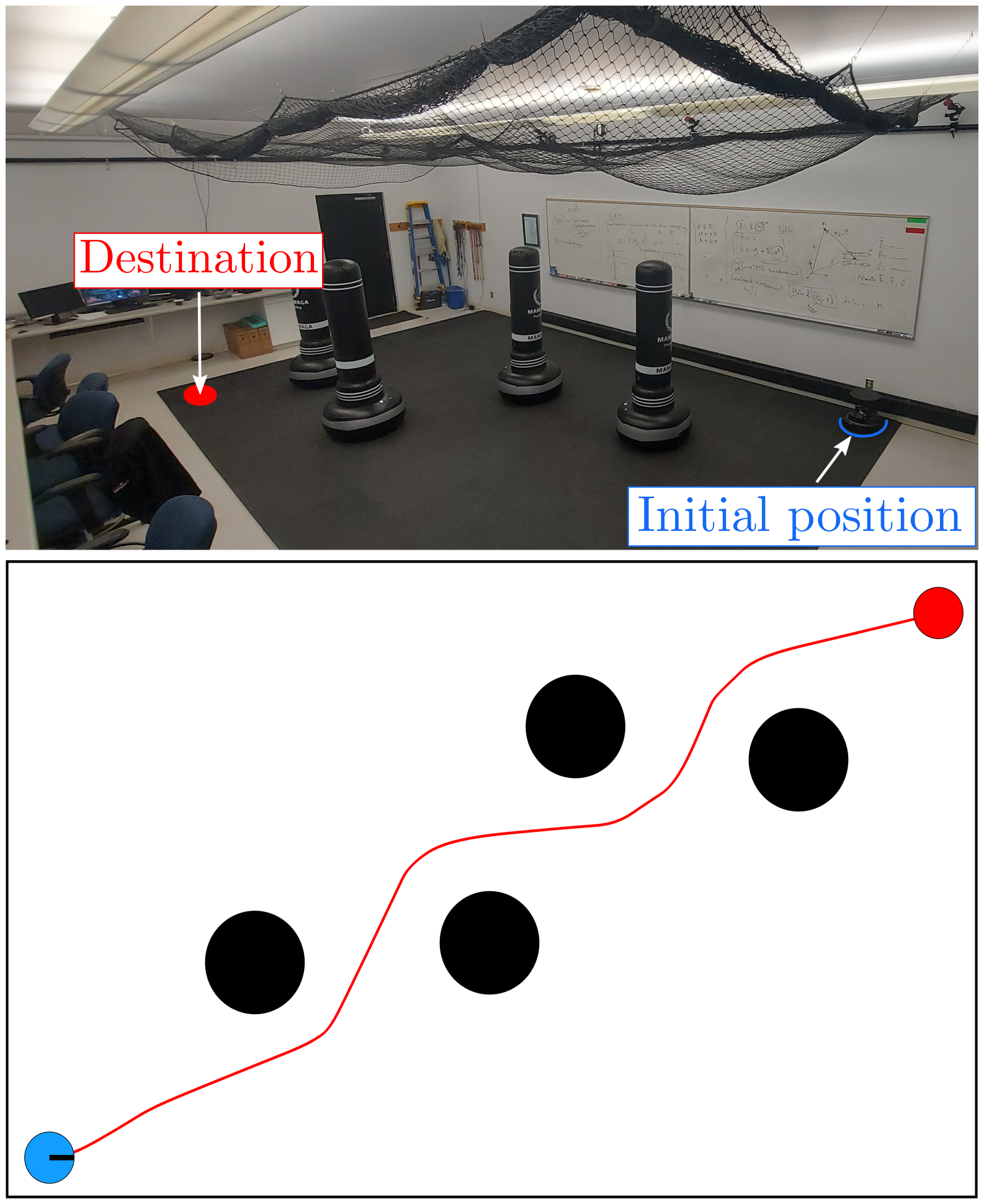}
\caption{The top figure shows the workspace configuration with initial and final positions highlighted. The bottom figure is a plot of the robot's trajectory in an orthographic projection of the workspace's top view. 
}
\label{fig:exp_results}
\end{figure}

We set up a $6.65m\times 4.2m$ workspace with four punching bags as obstacles. The robot is initially at the origin with its heading aligned with the $x$-axis ($\Phi=0$) of the workspace, and the target is set at the position $x_d = [6.1~3.6]^{\top}$. The experimental results are shown in Fig. \ref{fig:exp_results} and in a video that can be found online\footnote{\url{https://youtu.be/rQc062EDYts}}. The top figure of Fig. \ref{fig:exp_results} shows the workspace configuration with the initial and final positions. In the bottom figure, the trajectory of the robot is plotted in an orthographic projection top view of the workspace. The obtained results illustrate the safe navigation of the robot from the initial position to the final destination.

We also carried out a comparative experimental study with respect to some popular reactive feedback-based autonomous navigation algorithms, namely, the separating hyperplane approach (SH) \cite{Arslan2019}, the vector field histogram approach (VFH) \cite{VFH_91}, and the quasi-optimal navigation approach (QO) \cite{Ishak2023}. Under the same settings as in the previous experiment, the robot starts from the origin of the workspace with its heading aligned with the $x$-axis and navigates towards the target $x_d = [6.6 -3]$ as shown in Fig. \ref{fig:combined_results}-\subref{fig:compexp_results}. The generated trajectories with the four different algorithms are shown in Fig. \ref{fig:combined_results}-\subref{fig:compexp_results} and the video of the experiment can be found online\footnote{\url{https://youtu.be/KzUNLwQ5lMo}}. The execution time and path length of each algorithm are reported in Table \ref{fig:combined_results}-\subref{table_exp_time_comp}. One can notice that the quasi-optimal approach developed in our previous work \cite{Ishak2023} generates the shortest path and has the lowest execution time. Although the performance of our previously proposed quasi-optimal navigation approach is slightly better than that of the proposed hybrid feedback navigation approach, the former, unlike the latter, does not guarantee global asymptotic stability of the target location. Indeed, if the robot starts from the set of undesired equilibria, it can get stuck and have difficulty getting out, as shown in the video available online\footnote{\url{https://youtu.be/1gDqVkkAU0Y}}.
For the VFH approach, selecting an appropriate threshold\footnote{The lower bound of the polar histogram values, indicating the presence of obstacles.} depends on the workspace, which is crucial for safety and performance. In our experiment, the threshold was taken as 70000. The four implemented algorithms can be found online\footnote{\url{https://github.com/IshakChen9/navigate_TBT4_pkg.git}}.
A major limitation of our approach is that it currently applies only to environments with spherical obstacles. Nonetheless, it can be generalized to convex and even non-convex obstacles by employing conic hulls, which can be defined for any subset of $\mathbb{R}^n$ to characterize the shadow and {\it active regions}. Moreover, since shortest paths lie on tangents to obstacles \cite{ROHNERT198671}, the obstacle shape does not affect the optimality of the avoidance maneuver. A more challenging case arises when the robot starts inside cavities formed by non-convex obstacles, such as the cavity of a U-shaped obstacle, which can trap its motion. Fortunately, the hybrid systems framework used in our approach allows adding new modes ({\it e.g.,} a “Boundary following mode”) to address such scenarios. The reconstruction technique used in our sensor-based implementation is also restricted to spherical obstacles. Nevertheless, the sensor-based approach proposed in \cite{Ishak2023}, which applies local transformations to detected obstacle portions, provides a pathway to extend our local optimal avoidance maneuver to more general obstacle geometries.

\section{Conclusion}
In this work, an autonomous robot navigation scheme in $n$-dimensional Euclidean spaces with an arbitrary number of spherical obstacles was proposed. A hybrid feedback controller, switching between obstacle avoidance and motion-to-destination modes, ensures global navigation for a predefined destination. The proposed controller generates continuous control inputs and locally optimal obstacle avoidance maneuvers. Notably, the proposed scheme is implementable using only local sensor information, such as from LiDAR or vision systems. Experimental validation using the TurtleBot4 platform confirmed the effectiveness and practicality of the proposed approach. An extension of our approach to environments with arbitrarily shaped obstacles would be an interesting future work.   
\appendix
\subsection{Proof of Lemma \ref{lem2}}\label{appendix:Lem2}
The sets $\mathcal{F}$ and $\mathcal{J}$, defined in \eqref{sets}, are by construction closed subsets of $\mathbb{R}^n\times\mathbb{I}\times\mathbb{M}$, which shows that condition i) is satisfied.\\ 
Since the flow map $\mathrm{F}$, given in \eqref{flow_map},  is defined for all $\xi\in\mathcal{F}$, single-valued and continuous on $\mathcal{F}$, then $\mathcal{F}\subset\text{dom}(\mathrm{F})$, $\mathrm{F}$ is outer semicontinuous and bounded relative to $\mathcal{F}$, and convex for every $\xi\in\mathcal{F}$. Therefore, condition ii) is fulfilled.\\
The jump map $\mathrm{J}$, given in \eqref{jump_map}, is single-valued on $\mathcal{J}_m\times\{m\}$, $m\in\{-1,1\}$. Since the angles $\varphi_k^1,\varphi_k^{-1}$ are selected according to Lemma \ref{lem1}, we have $\mathcal{C}_{\mathbb{R}^n}^{\leq}(c_k,v_k^1,\varphi_k^1)\cap\mathcal{C}_{\mathbb{R}^n}^{\leq}(c_k,v_k^{-1},\varphi_k^{-1})=\{c_k\}$, which implies that $\mathcal{C}_{\mathbb{R}^n}^{\geq}(c_k,v_k^1,\varphi_k^1){\cup}\mathcal{C}_{\mathbb{R}^n}^{\geq}(c_k,v_k^{-1},\varphi_k^{-1})=\mathbb{R}^n$. Therefore, given \eqref{jump_map_B}, $B(x,k)\neq\varnothing$ for all $(x,k)\in\mathbb{R}^n\times\mathbb{I}$, then $M(\xi)\neq\varnothing$ for all $\xi\in\mathcal{J}_0\times\{0\}$, and hence $\mathrm{J}(\xi)\neq\varnothing$ for all $\xi\in\mathcal{J}_0\times\{0\}$. Thus, $\mathcal{J}\in\text{dom}(\mathrm{J})$. Moreover, $\mathrm{J}$ has a closed graph relative to $\mathcal{J}_0\times\{0\}$ as $B$ is allowed to be set-valued whenever $x\in\cap_{m=-1,1}\mathcal{C}_{\mathbb{R}^n}^{\leq}(c_k,v_k^m,\varphi_k^m)$. Then,
according to \cite[Lemma 5.10]{Sanfelice}, $\mathrm{J}$ is outer semicontinuous relative to $\mathcal{J}$. Furthermore, the jump map $\mathrm{J}$ is locally
bounded relative to $\mathcal{J}$ since $M$ and $K$ take values over finite
discrete sets $\mathbb{I}$ and $\mathbb{M}$, and the remaining component of $\mathrm{J}$ is a single-valued continuous function on $\mathcal{J}$, which shows the satisfaction of condition iii) and completes the proof.
\subsection{Proof of Theorem \ref{the1}}\label{appendix:the1}
\textbf{Item i):} First, we prove that $\mathcal{F}\cup\mathcal{J}=\mathcal{K}$, which boils down to show that for each $k\in\mathbb{I}$, and $m\in\{-1,1\}$,
\begin{align}\label{safety_cdt}
\Tilde{\mathcal{F}}_0\cup\Tilde{\mathcal{J}}_0=\mathcal{F}_k^m\cup\mathcal{J}_k^m=\mathcal{X},
\end{align}
since the satisfaction of \eqref{safety_cdt}, along with \eqref{set_0}, \eqref{set_m}, and \eqref{sets}, implies that $\mathcal{F}\cup\mathcal{J}=\mathcal{X}\times\mathbb{I}\times\mathbb{M}=\mathcal{K}$. We start by showing $\Tilde{\mathcal{F}}_0\cup\Tilde{\mathcal{J}}_0=\mathcal{X}$ in \eqref{safety_cdt}. Recall that $\Tilde{\mathcal{F}}_0:=\cap_{k\in\mathbb{I}}\mathcal{F}_k^0$, $\Tilde{\mathcal{J}}_0:=\cup_{k\in\mathbb{I}}\mathcal{J}_k^0$, and for each $k\in\mathbb{I}$, $\mathcal{F}_k^0:=\overline{\mathcal{X}\setminus\mathcal{A}_k(x_d)},\;\mathcal{J}_k^0:=\mathcal{A}_k(x_d)$, as defined in \eqref{sets_m0}, and hence,
\[
\small
\begin{array}{l}

    \left(\bigcap\limits_{k\in\mathbb{I}}\mathcal{F}_k^0 \right)\cup\left(\bigcup\limits_{k\in\mathbb{I}}\mathcal{J}_k^0 \right)= \left(\bigcap\limits_{k\in\mathbb{I}}\overline{\mathcal{X}\setminus\mathcal{A}_k(x_d)} \right)\cup\left(\bigcup\limits_{k\in\mathbb{I}}\mathcal{A}_k(x_d) \right)\\
    =\bigcap\limits_{k\in\mathbb{I}}\left\{\overline{\mathcal{X}\setminus\mathcal{A}_k(x_d)} \cup\left(\bigcup\limits_{i\in\mathbb{I}}\mathcal{A}_i(x_d) \right)\right\}=\bigcap\limits_{k\in\mathbb{I}}\overline{\mathcal{X}}=\mathcal{X},
\end{array}
\]
where $\mathcal{X}$ is a closed set as defined in \eqref{freespace}. Now we prove that for each $k\in\mathbb{I}$, and $m\in\{-1,1\}$, $\mathcal{F}_k^m\cup\mathcal{J}_k^m=\mathcal{X}$. Note that from \eqref{sets_11}, $\mathcal{F}_k^m:=\mathcal{A}_k(x_k^{m})\setminus\mathcal{C}_{\mathcal{X}}^{<}(c_k,v_k^m,\varphi_k^m),\mathcal{J}_k^m:=\overline{\mathcal{X}\setminus\mathcal{F}_k^m}$, and thus, $\mathcal{F}_k^m\cup\mathcal{J}_k^m=\mathcal{F}_k^m\cup(\overline{\mathcal{X}\setminus\mathcal{F}_k^m})=\mathcal{X}$.\\
Let us define the set $\mathcal{M}_{\mathcal{H}}(\mathcal{K})$ of all maximal solutions to the hybrid system \eqref{Cl_ctrl} represented by its data $\mathcal{H}$ with $\xi(0,0)\in\mathcal{K}$. Each solution $\xi \in\mathcal{M}_{\mathcal{H}}(\mathcal{K})$ has range $\text{rge}\,\xi \subset\mathcal{K}=\mathcal{F}\cup\mathcal{J}$. The augmented state space $\mathcal{K}$ is forward invariant for $\mathcal{H}$ if, for each $\xi(0,0) \in \mathcal{K}$, there exists one solution, and every $\xi\in \mathcal{M}_{\mathcal{H}}(\mathcal{K})$ is complete and has range $\text{rge}\,\xi \subset\mathcal{K}$ as per \cite[Definition~3.3]{Forward_invariance_sanfelice}. The forward invariance of $\mathcal{K}$ is then shown by the c=ompleteness of the solutions $\xi\in\mathcal{M}_{\mathcal{H}}(\mathcal{K})$ which we prove using \cite[Proposition~6.10]{Sanfelice}. We start by showing the following viability condition
\begin{align}\label{viability_cdt}
    \mathrm{F}(\xi)\cap\mathrm{T}_{\mathcal{F}}(\xi)\neq\varnothing, \forall \xi\in\mathcal{F}\setminus\mathcal{J},
\end{align}
where $\mathrm{T}_{\mathcal{F}}(\xi)$ is Bouligand's tangent cone of the set $\mathcal{F}$ at $\xi$ as defined in \cite[Definition~5.12]{Sanfelice}. Inspired by \cite[Appendix~1]{SoulaimaneHybTr}, we proceed as follows. Let $\xi=(x,k,m)\in\mathcal{F}\setminus\mathcal{J}$, which implies by \eqref{sets} that $(x,k)\in\mathcal{F}_m\setminus\mathcal
{J}_m$ for some $m\in\mathbb{M}$. Consider the two cases (modes) $m=0$ and $m\in\{-1,1\}$. For $m=0$, as per definition \eqref{set_0}, there exists $k\in\mathbb{I}$ such that $x\in\Tilde{\mathcal{F}}_0\setminus\Tilde{\mathcal{J}}_0$. When $x\in\mathring{\Tilde{\mathcal{F}}}_0\setminus\Tilde{\mathcal{J}}_0$, then $x$ is in the interior of the set $\Tilde{\mathcal{F}}_0$, and hence, $\mathrm{T}_{\mathcal{F}}(\xi)=\mathbb{R}^n\times\{0\}\times\{0\}$ and \eqref{viability_cdt} holds. When $x\in\partial\Tilde{\mathcal{F}}_0\setminus\Tilde{\mathcal{J}}_0$, it is clear, according to \eqref{active_region} and \eqref{sets_m0}, that $x$ must be on the boundary of one of the obstacles and does not belong to the {\it active region} (\textit {i.e.,} $\partial\Tilde{\mathcal{F}}_0\setminus\Tilde{\mathcal{J}}_0\subseteq\bigcup_{k\in\mathbb{I}}(\partial\mathcal{O}_k\setminus\mathcal{A}_k(x_d))$. Then, $\mathrm{T}_{\mathcal{F}}(x,k,0)=\mathcal{P}_{\geq}(x,x-c_k)\times\{0\}\times\{0\}$. Since $u(x,k,0)=-\gamma(x-x_d)$, $u(x,k,0)^{\top}(x-c_k)>0$ and \eqref{viability_cdt} holds. Now, when $m\in\{-1,1\}$, according to \eqref{sets_11}, there exists $k\in\mathbb{I}$ such that $x\in\mathring{\mathcal{F}}^k_m\setminus\mathcal{J}_k^m$. For $x\in\mathring{\mathcal{F}}^k_m\setminus\mathcal{J}_k^m$, $\mathrm{T}_{\mathcal{F}}(x,k,m)=\mathbb{R}^n\times\{0\}\times\{0\}$ and \eqref{viability_cdt} holds. When $x\in\partial\mathcal{F}_k^m\setminus\mathcal{J}_k^m$, it is clear, according to \eqref{sets_11}, that $x$ must be on the boundary of obstacle $\mathcal{O}_k$ and belongs to the {\it active region} (\textit{i.e.,} $\partial\mathcal{F}_k^m\setminus\mathcal{J}_k^m\subseteq\partial\mathcal{O}_k\cap\mathcal{A}_k(x_k^m)$). Thus, $\mathrm{T}_{\mathcal{F}}(x,k,m)=\mathcal{P}_{\geq}(x,x-c_k)\times\{0\}\times\{0\}$. From \eqref{ctrl_m11} and \eqref{hyb_ctrl_L}, and since $x\in\partial\mathcal{O}_k$, then $\theta(x,k)=\frac{\pi}{2}$ and $u(x,k,m)=-\gamma\mu(x,k,m)\left(x_k^m-x-\|x_k^m-x\|\cos(\beta(x,k,m))\frac{c_k-x}{\|c_k-x\|}\right)$ where $\beta(x,k,m)=\angle(c_k-x,x_k^m-x)$ and $\mu(x,k,m)=$ $\left(1+\frac{e_k}{\|x-x_k^m\|}\frac{\beta(x,k,m)}{\theta(x,k)}\right)>0$. Hence,
{\allowdisplaybreaks
\small
\begin{align*}
&u(x,k,m)^{\top}(x-c_k)=\gamma\mu(x,k,m)((x_k^m-x)^{\top}(x-c_k)\\
&\qquad-\frac{\|x_k^m-x\|}{\|c_k-x\|}\cos(\beta(x,k,m))(c_k-x)^{\top}(x-c_k))\\
&=\gamma\mu(x,k,m)((x_k^m-x)^{\top}(x-c_k)\\
&\qquad+\|x_k^m-x\|\|c_k-x\|\cos(\beta(x,k,m))\\
&=\gamma\mu(x,k,m)((x_k^m-x)^{\top}(x-c_k)+(x_k^m-x)^{\top}(c_k-x))=0,
\end{align*}
}
and \eqref{viability_cdt} holds for $m\in\{-1,1\}$. Therefore, according to \cite[Proposition 6.10]{Sanfelice}, since \eqref{viability_cdt} holds
for all $\xi\in\mathcal{F}\setminus\mathcal{J}$, there exists a nontrivial solution to the hybrid system $\mathcal{H}$ for
every initial condition in $\mathcal{K}$. Finite escape times can only occur via the flow. They cannot occur for $x\in\mathcal{F}_k^{-1}\cup\mathcal{F}_k^1$,  as the sets $\mathcal{F}_k^{-1}$ and $\mathcal{F}_k^1$ are bounded by their definition \ref{sets_11}, nor for $x\in\Tilde{\mathcal{F}}_0$ since they would make $(x-x_d)\top(x-x_d)$ grow unbounded, which would contradict the fact that $\frac{d}{dt}((x-x_d)\top(x-x_d))\leq0$ by the definition of $u(x,k,0)$. 
Therefore, all maximal solutions do not have finite escape times. Moreover, according to \eqref{Cl_ctrl},
$x^+ = x$, and from the definitions \eqref{jump_map_M},
\eqref{jump_map_B}, and \eqref{jump_map}, it follows that $\mathrm{J}(\mathcal{J})\subset \mathcal{K}$. Thus, solutions of the hybrid system \eqref{Cl_ctrl} cannot leave $\mathcal{K}$ through jumps and, as per \cite[Proposition 6.10]{Sanfelice}, all maximal solutions
are complete.\\
 \textbf{Item ii):} Using \cite[Definition~7.1]{Sanfelice}, we first show the stability of $\mathcal{A}$, and then its attractivity assuming Zeno-free solutions. Since $x_d\in\mathring{\mathcal{X}}$, there exists $\bar{\varepsilon}>0$ such that $\mathcal{B}(x_d,\bar{\varepsilon})\cap\mathcal{O}_k=\varnothing,\,\forall k\in\mathbb{I}$. As per the sets definitions in \eqref{sets_m0} and \eqref{set_0}, $\mathcal{B}(x_d,\varepsilon)\subset\Tilde{\mathcal{F}}_0$ for all $\varepsilon\in[0,\bar{\varepsilon}]$. Thus, $\mathcal{B}(x_d,\varepsilon)\cap\Tilde{\mathcal{J}}_0=\varnothing$, and $x$ evolves under $\Dot{x}=-\gamma(x-x_d)$, which implies forward invariance of the set $\mathcal{B}:=\mathcal{B}(x_d,\varepsilon)\times\mathbb{I}\times\mathbb{M}$. Therefore, according to \cite[Definition~7.1]{Sanfelice}, the set $\mathcal{A}$ is stable for the hybrid system \eqref{Cl_ctrl}. Now, let us prove the attractivity of $\mathcal{A}$, but first, we need the following lemma.
\begin{lemma}\label{lem8}
    For a given obstacle index $k\in\mathbb{I}$, the {\it obstacle-avoidance} mode $m\in\{-1,1\}$ activates only once and remains active for a finite hybrid time interval $[t_0^{k,m},t_f^{k,m}]\times[j_0^{k,m},j_f^{k,m}]$ where $(t_0^{k,m},j_0^{k,m})$ and $(t_f^{k,m},j_f^{k,m})$ are, respectively, the activation and deactivation hybrid times of mode $m$ for the obstacle with index $k$.
\end{lemma}
{\it Proof:} Consider an obstacle index $k\in\mathbb{I}$ and the positive definite function $V_m(x)=\frac{1}{2}||x-x_k^m||^2$, for $m\in\{-1,1\}$ and $x\in\mathcal{F}_k^m$. The time derivative of $V_m(x)$ is given by
{\allowdisplaybreaks
\small
\begin{align*}
    &\Dot{V}_m(x)=\frac{\partial V_m(x)}{\partial x}^{\top}\Dot{x}\\
 &=(x-x_k^m)^\top \big(\alpha(x,k)\mu(x,k,m)\kappa(x,k,m)\\
 &\quad+(1-\alpha(x,k))u_d(x)\big),\\
    &=-(x_k^m-x)^{\top}\bigg\{K_1\bigg(\gamma(x_k^m-x)-\tau(x,k,m)\frac{c_k-x}{\|c_k-x\|}\bigg)\\
    &\quad+\gamma K_2(x_d-x)\bigg\},\\
&=-\bigg\{K_1\bigg(\gamma\|x_k^m-x\|^2-\tau(x,k,m)\frac{(x_k^m-x)^\top(c_k-x)}{\|c_k-x\|}\bigg)\\
&\quad+\gamma K_2(x_k^m-x)^\top(x_d-x)\bigg\},\\
   &=-\gamma\bigg\{K_1\bigg(\|x_k^m-x\|^2\\
   &\quad-\frac{\|x_k^m-x\|}{\|c_k-x\|}\frac{\sin(\theta(x,k)-\beta(x,k,m))}{\sin(\theta(x,k))}(x_k^m-x)^\top(c_k-x)\bigg)\\
   &\quad+K_2(x_k^m-x)^\top(x_d-x)\bigg\},\\
    &=-\gamma\bigg\{K_1\bigg(\|x_k^m-x\|^2\\
    &\quad-\|x_k^m-x\|^2\frac{\sin(\theta(x,k)-\beta(x,k,m))}{\sin(\theta(x,k))}\cos(\beta(x,k,m))\bigg)\\
    &\quad+K_2(x_k^m-x)^\top(x_d-x)\bigg\},\\
    &=-\gamma\bigg(K_1\|x_k^m-x\|^2\frac{\sin(\beta(x,k,m))}{\sin(\theta(x,k))}\cos(\theta(x,k)-\beta(x,k,m))\\
    &\quad+K_2(x_k^m-x)^\top(x_d-x)\bigg),
\end{align*}}
where $K_1=\alpha(x,k)\mu(x,k,m)\geq0$, $K_2=1-\alpha(x,k)\geq0$, and we used the fact that $\sin(\theta(x,k))-\sin(\theta(x,k)-\beta(x,k,m))\cos(\beta(x,k,m))=\sin(\beta(x,k,m))\cos(\theta(x,k)-\beta(x,k,m))$. Since $0<\theta(x,k)\leq\frac{\pi}{2}$ and $0\leq\beta(x,k,m)\leq\theta(x,k)$, one can deduce that
\begin{align}\label{eq1}
    K_1\|x_k^m-x\|^2\frac{\sin(\beta(x,k,m))}{\sin(\theta(x,k))}\cos(\theta(x,k)-\beta(x,k,m))\geq 0,
\end{align}
for all $(x,k,m)\in\mathcal{F}_m\times\{-1,1\}$. Next, we prove that $(x_k^m-x)^\top(x_d-x)\geq0$. The following fact will be needed in the rest of the proof.
\begin{fact}\label{fact1}
    $\forall (k,m)\in\mathbb{I}\times\{-1,1\},\; \mathcal{F}_k^m\subset\mathcal{P}_{\leq}(p_k,x_d-p_k)$. 
\end{fact}
{\it Proof:} Since $p_k\in\partial\mathcal{O}_k$ and $x_k^m\in\mathcal{P}_{\leq}(p_k,c_k-p_k)$ \big({\it i.e., $(c_k-p_k)^\top(x_k^m-p_k)\leq0$}\big), then $p_k\in\mathcal{C}_{\mathcal{X}}^{\leq}(x_k^m,c_k-x_k^m,\theta_k(x_k^m))\setminus\mathcal{S}_k(x_k^m)$. Therefore, $\mathcal{S}_k(x_k^m)\subset\mathcal{P}_{\geq}(p_k,c_k-p_k)$. Moreover, since $\mathcal{F}_k^m\subset\mathcal{S}_k(x_k^m)$ (see definitions \eqref{sets}, \eqref{active_region}) and $p_k=c_k+r_k(x_d-c_k)/\|x_d-c_k\|$, one can deduce that $\mathcal{F}_k^m\subset\mathcal{P}_{\leq}(p_k,x_d-p_k)$ for $(k,m)\in\mathbb{I}\times\{-1,1\}$, which concludes the proof of Fact \ref{fact1}.\\
According to \eqref{set_m}, $(x,k)\in\mathcal{F}_m$ implies that $x\in\mathcal{F}_k^m$, and hence, according to \eqref{sets_11}, 
\begin{align}\label{prop1}
    (c_k-x)^\top(x_k^m-x)\geq0.
\end{align}
Moreover, since $x_k^m\in\mathcal{H}_k(x_d)\cap\mathcal{P}_{\geq}(p_k,x_d-p_k)$, one has
{\small
\begin{subequations}
\begin{align}
    &(x_d-p_k)^\top(x_k^m-p_k)\geq0,\\
    &\left(x_d-c_k-r_k\frac{x_d-c_k}{\|x_d-c_k\|}\right)^\top\left(x_k^m-c_k-r_k\frac{x_d-c_k}{\|x_d-c_k\|}\right)\geq0,\\
    &\left(1-\frac{r_k}{\|x_d-c_k\|}\right)(x_d-c_k)^\top\left(x_k^m-c_k-r_k\frac{x_d-c_k}{\|x_d-c_k\|}\right)\geq0,\\
    &(x_d-c_k)^\top(x_k^m-c_k)-r_k\|x_d-c_k\|\geq0,\\
    &(x_d-c_k)^\top(x_k^m-c_k)\geq r_k\|x_d-c_k\|.\label{prop2}
\end{align}
\end{subequations}}
In addition, as per Fact \ref{fact1}, for all $x\in\mathcal{F}_k^m$, one has
{\small
\begin{subequations}
\begin{align}
    &(x_d-p_k)^\top(x-p_k)\leq0,\\
    &\left(x_d-c_k-r_k\frac{x_d-c_k}{\|x_d-c_k\|}\right)^\top\left(x-c_k-r_k\frac{x_d-c_k}{\|x_d-c_k\|}\right)\leq0,\\
    &\left(1-\frac{r_k}{\|x_d-c_k\|}\right)(x_d-c_k)^\top\left(x-c_k-r_k\frac{x_d-c_k}{\|x_d-c_k\|}\right)\leq0,\\
    &(x_d-c_k)^\top(x-c_k)-r_k\|x_d-c_k\|\leq0,\\
    &(x_d-c_k)^\top(x-c_k)\leq r_k\|x_d-c_k\|.\label{prop3}
\end{align}
\end{subequations}}
From \eqref{prop2} and \eqref{prop3}, one gets
\begin{align*}
    (x_d-c_k)^\top(x_k^m-c_k)-(x_d-c_k)^\top(x-c_k)\geq0,\\
    (x_d-c_k)^\top(x_k^m-x)\geq0,\\
    (x_d-x)^\top(x_k^m-x)+(x-c_k)^\top(x_k^m-x)\geq0,\\
    (x_d-x)^\top(x_k^m-x)\geq (c_k-x)^\top(x_k^m-x),
\end{align*}
and according to \eqref{prop1}, one can show that $(x_d-x)^\top(x_k^m-x)\geq0.$
Therefore, $\Dot{V}_m(x)\leq0$ for all $(x,k,m)\in\mathcal{F}_k^m\times\mathbb{I}\times\{-1,1\}$. Moreover, $\Dot{V}_m(x)=0$ only if $x\in\mathcal{L}_k(x_k^m)$, which is excluded from the set $\mathcal{F}_k^m$ for $m\in\{-1,1\}$, then one can conclude that $V_m(x)<0$ for all $(x,k,m)\in\mathcal{F}_k^m\times\mathbb{I}\times\{-1,1\}$. Thus, as $x_k^m\notin\mathcal{F}_k^m$, $k\in\mathbb{I}$, there exists a hybrid time $t_f^{k,m}>0$, $j_f^{k,m}\in\mathbb{N}\setminus\{0\}$,  such that $x(t_f^{k,m},j_f^{k,m})$ leaves $\mathcal{F}_k^m$ and the mode $m(j_f^{k,m},j_f^{k,m})$ jumps to the {\it motion-to-destination} mode. Therefore, if the {\it obstacle-avoidance} mode $\left(m\in\{-1,1\}\right)$ is activated at $(t_0^{k,m},j^{k,m}_0)$, the hybrid time interval for this mode $m$ is given by $[t_0^{k,m},t_f^{k,m}]\times[j_0^{k,m},j_f^{k,m}]$. We now show that the {\it obstacle-avoidance} mode can be activated at most once for any given obstacle $\mathcal{O}_k,\,k\in\mathbb{I}$. Since the direction of the control \eqref{ctrl_m11} in the {\it obstacle-avoidance} mode ($m=\pm1$) is tangent to the obstacle, the robot exits the flow set $\mathcal{F}_k^m$ from the region $\partial\mathcal{O}_k\cap\mathcal{E}(x_k^m)$ and switches to the {\it motion-to-destination} mode. On the other hand, under the control $u_d(x)=-\gamma(x-x_d)$ of the {\it motion-to-destination} mode, the robot can only enter its jump set $\mathcal{J}_k^0$ from the lower boundary ({\it i.e.,} $\partial\mathcal{J}_k^0\cap\partial\mathcal{B}(c_k,\bar{r}_k)$) to switch to the {\it obstacle-avoidance} mode. Furthermore, since the distance to the target in the {\it motion-to-destination} mode always decreases, it follows that once an obstacle is avoided, it cannot be encountered again, which concludes the proof of Lemma \ref{lem8}.\\

Lemma \ref{lem8} shows that the {\it obstacle-avoidance} mode can be activated at most once for any given obstacle and remains active for a finite hybrid time interval $[t_0^{k,m},t_f^{k,m}]\times[j_0^{k,m},j_f^{k,m}]$. The mode then jumps to the {\it motion-to-destination} mode ({\it i.e.,} $m=0$) and $x$ leaves the flow set $\mathcal{F}_k^m$ to the flow set $\mathcal{F}_k^0$ of the {\it motion-to-destination} mode. Therefore, since the number of obstacles is finite, there exists a finite sequence of obstacles to avoid, and after the last avoidance in this sequence, the mode $m$ jumps to the {\it motion-to-destination} mode where the flow $\Dot{x}=-\gamma(x-x_d)$ guarantees the attractivity of the equilibrium set $\mathcal{A}$.
\subsection{Proof of Lemma \ref{lem6}}\label{appendix:Lem6}
For the hybrid time interval $[t_0^k,t_f^k]\times[j_0^k,j_f^k]$ where obstacle $k$ is selected for avoidance, the robot, according to the jump maps $K(\cdot)$ and $M(\cdot)$ defined in \eqref{jump_map_K}-\eqref{jump_map_M}, has to first operate in the {\it obstacle-avoidance} mode ($m=\pm1$) and then in the {\it motion-to-destination} mode ($m=0$). Let $(t^k_s,j^k_s)$ be the hybrid time at which the mode selector jumps from $m=\pm1$ to $m=0$. Let us consider the first case where $x(t_0^k,j_0^k)\in\mathcal{J}_0^k\setminus\mathcal{L}(x_d,c_k)$. In the first mode ($m\pm 1$), the velocity vector $u(x,k,m)$, defined in \eqref{ctrl_m11}, is a function of three vectors $(c_k-x)$, $(x_d-x)$, and $(x_k^m-x)$. Thus, if $x_k^m\in\mathcal{H}_k(x_d)\cap\mathcal{P}_{\geq}(p_k,x_d-p_k)\cap\mathcal{PL}\left (x_d,c_k,x(t_0^k,j_0^k)\right)$, then the points $x_d,c_k,x(t_0^k,j_0^k),x_k^m$ are contained in the two-dimensional plane $\mathcal{PL}\left(x_d,c_k,x(t_0^k,j_0^k)\right)$, and the vectors $(c_k-x(t_0^k,j_0^k))$ and $(x_d-x(t_0^k,j_0^k))$ are linearly independent. Therefore, $x(t,j)\in\mathcal{PL}\left(x_d,c_k,x(t_0^k,j_0^k)\right)$ for all $(t,j)\in[t_0^k,t_s^k]\times[j_0^k,j_s^k]$. In the second mode ($m=0$), the velocity vector is given by $u(x,k,0)=\gamma(x_d-x)$, which implies that the resultant trajectory is the line segment joining $x(t_s^k,j_s^k)$ and $x(t_f^k,j_f^k)$ where $\left(x(t_f^k,j_f^k)-x(t_s^k,j_s^k)\right)$ and $\left(x_d-x(t_s^k,j_s^k)\right)$ are collinear. Therefore, since $x_d,x(t_s^k,j_s^k)\in\mathcal{PL}\left(x_d,c_k,x(t_0^k,j_0^k)\right)$, as shown in the previous mode, then $x(t,j)\in\mathcal{PL}\left(x_d,c_k,x(t_0^k,j_0^k)\right)$ for all $(t,j)\in[t_s^k,t_f^k]\times[j_s^k,j_f^k]$. Finally, one can conclude that $x(t,j)\in\mathcal{PL}\left(x_d,c_k,x(t_0^k,j_0^k)\right)$ for all $(t,j)\in[t_0^k,t_f^k]\times[j_0^k,j_f^k]$. In the second case where $x(t_0^k,j_0^k)\in\mathcal{J}_0^k\cap\mathcal{L}(x_d,c_k)$, the points $x_d,c_k,x(t_0^k,j_0^k)$ are aligned. Therefore, we take any point $y$ of the free space not aligned with points $x_d,c_k$ $\big(${\it i.e.,} $y\in\mathbb{R}^n\setminus\mathcal{L}(x_d,c_k)\big)$ so that we can consider the two-dimensional plane $\mathcal{PL}(x_d,c_k,y)$ and select the virtual destination $x_k^m\in\mathcal{H}_k(x_d)\cap\mathcal{P}_{\geq}(p_k,x_d-p_k)\cap\mathcal{PL}(x_d,c_k,y)$. The proof becomes similar to that of the first case when considering the plane $\mathcal{PL}(x_d,c_k,y)$.
\subsection{Proof of Proposition \ref{proposition1}}\label{appendix:proposition1}
We first start by proving the continuity of the velocity control law $u(x,k,m)$. Since $u(x,k,m)$ is continuous during the flow, we only need to verify its continuity at the switching instances, which corresponds to jumps between the two modes, the {\it motion-to-destination} mode ($m=0$), and the {\it obstacle-avoidance} mode ($m=\pm1$).\\
\textbf{Case 1} ($m=0\rightarrow\pm1$): The jump from the {\it motion-to-destination} mode to the {\it obstacle-avoidance} mode occurs when the robot enters the jump set $\mathcal{J}_k^0$ through the boundary $\big(\partial\mathcal{J}_k^0\cap\partial\mathcal{B}(c_k,\Bar{r}_k)\big)$. The control law for both modes is given by $u(x,k,0)=\gamma(x_d-x)$ and $u(x,k,m)=\alpha(x,k)\mu(x,k,m)u_{m}(x,k)+\gamma(1-\alpha(x,k))(x_d-x)$ for $m=\pm1$. Since, for all $x\in\partial\mathcal{J}_k^0\cap\partial\mathcal{B}(c_k,\Bar{r}_k)$, $d(x,\mathcal{O}_k)=\Bar{r}_k$, then $\alpha(x,k)=0$ and $u(x,k,m)=u(x,k,0)$.\\
\textbf{Case 2} ($m=\pm1\rightarrow0$): Let $\mathcal{O}_k,\,k\in\mathbb{I},$ be the obstacle selected for the avoidance. Since the virtual destinations are selected as in Lemma \ref{lem6}, the motion during {\it obstacle-avoidance} and {\it motion-to-destination} modes, while obstacle $\mathcal{O}_k$ is selected, is two-dimensional. The motion takes place on the plane $\mathcal{PL}\left(x_d,c_k,x(t_0^k,j_0^k)\right)$ if $x(t_0^k,j_0^k)\notin\mathcal{L}(x_d,c_k)$ $\bigl($resp. $\mathcal{PL}(x_d,c_k,y)$, where $y\in\mathbb{R}^n\setminus\mathcal{L}(x_d,c_k)$, if $x(t_0^k,j_0^k)\in\mathcal{L}(x_d,c_k)\bigr)$ where $(t_0,j_0)$ is the hybrid time at which the mode variable $m$ jumps to the {\it obstacle-avoidance} mode. Therefore, the navigation problem, while obstacle $\mathcal{O}_k$ is selected, is reduced to the two-dimensional case. According to the jump maps defined in \eqref{jump_map_M}-\eqref{jump_map_B_hat}, the mode variable $m$ jumps to $1$ when $x\in\bigl(\mathcal{J}_k^0\cap\mathcal{C}_k^1\bigr)\cup\bigl(\mathcal{J}_k^0\cap\mathcal{C}_k\cap\mathcal{P}_{<}(c_k,x_k^{-1}-x_k^1)\bigr)=\mathcal{J}_k^0\cap\mathcal{P}_{<}(c_k,x_k^{-1}-x_k^1)$.
Similarly, $m$ jumps to $-1$ when $x\in\mathcal{J}_k^0\cap\mathcal{P}_{>}(c_k,x_k^{-1}-x_k^1)$. When $x\in\mathcal{J}_k^0\cap\mathcal{P}_{=}(c_k,x_k^{-1}-x_k^1)$, $m$ can jump to $1$ or $-1$. Thus, one can deduce that the variable $m$ always jumps to the {\it obstacle-avoidance} mode in which the associated virtual destination is the closest to $x$. Furthermore, as the virtual destinations belong to the cone {\it hat} $\mathcal{H}_k(x_d)$ of vertex $x_d$ enclosing obstacle $\mathcal{O}_k$, the jump will occur when $x\in\mathcal{E}(x_d)\cap\mathcal{E}(x_k^m)\cap\mathcal{PL}\left(x_d,c_k,x(t_0^k,j_0^k)\right)$ if $x(t_0^k,j_0^k)\notin\mathcal{L}(x_d,c_k)$ $\bigl($resp. $x\in\mathcal{E}(x_d)\cap\mathcal{E}(x_k^m)\cap\mathcal{PL}(x_d,c_k,y)$, where $y\in\mathbb{R}^n\setminus\mathcal{L}(x_d,c_k)$, if $x(t_0^k,j_0^k)\in\mathcal{L}(x_d,c_k)\bigr)$ for $~m=~\pm1$.\\
Thus, $\frac{(x_d-x)^\top}{\|x_d-x\|}\frac{(x_k^m-x)}{\|x_k^m-x\|}=1$ and $\|x_d-x\|=\|x_k^m-x\|+\|x_d-x_k^m\|$ ({\it i.e.,} the points $x,x_k^m$, and $x_d$ are aligned). In addition, $\beta(x,k,m)=\angle(c_k-x,x_k^m-x)=\theta(x,k)$ when $x\in\mathcal{E}(x_k^m)$. Therefore, $\tau(x,k,m)=0$ and $\mu(x,k,m)=\frac{\|x-x_k^m\|+e_k}{\|x-x_k^m\|}$. Hence, $u(x,k,m)=\gamma\alpha(x,k)\frac{\|x-x_k^m\|+e_k}{\|x-x_k^m\|}(x_k^m-x)+\gamma(1-$ $\alpha(x,k))(x_d-x)$, and since $e_k=\|x_d-x_k^m\|$, one has $u(x,k,m)=\gamma\alpha(x,k)\|x_d-x\|\frac{x_k^m-x}{\|x_k^m-x\|}+\gamma(1-\alpha(x,k))(x_d-x)=\gamma\alpha(x,k)(x_d-x)+\gamma(1-\alpha(x,k))(x_d-x)=\gamma(x_d-x)=u(x,k,0)$.\\
Now we prove that the obstacles are avoided through local optimal obstacle-avoidance maneuvers. We use the result of \cite[Lemma 1]{Ishak2023}, stating that, in the case of a single spherical obstacle, the shortest path is obtained if the obstacle-avoidance maneuver is optimal ({\it i.e.,} the velocity vector is tangent to the obstacle and minimizes the deviation with respect to the nominal control $u_d(x)$ in the {\it shadow region}) and the motion-to-destination is performed under the nominal control $u_d(x)$. According to Lemma \ref{lem6}, the control $\kappa(x,k,m)$ of the {\it obstacle-avoidance} mode, defined in \eqref{ctrl_m11}, satisfies the optimality conditions of the obstacle-avoidance maneuver with respect to a given virtual destination $x_k^m$ and a given obstacle $\mathcal{O}_k$ in the {\it active region} $\mathcal{A}_k$ where $m\in\{-1,1\}$ and $k\in\mathbb{I}$. Since $u(x,k,m)=\alpha(x,k)\mu(x,k,m)\kappa(x,k,m)+(1-\alpha(x,k))u_d(x)$ for $m=\pm1$, and $\alpha=1$ when $d(x,\mathcal{O}_k)\leq\bar{r}_k-\epsilon$, $u(x,k,m)=\mu(x,k,m)\kappa(x,k,m)$ where $\mu(x,k,m)$ is a positive scalar function. Therefore, the hybrid control law $u(x,k,m)$ satisfies the optimality conditions of the obstacle-avoidance maneuver with respect to a given virtual destination $x_k^m$ and a given obstacle $\mathcal{O}_k$ for all $x\in\mathcal{F}_k^m\cap\mathcal{B}(c_k,r_k+\bar{r}_k-\epsilon)$, $m\in\{-1,1\}$ and $k\in\mathbb{I}$. Now we show that the optimality conditions of the obstacle-avoidance maneuver are also satisfied with respect to the destination $x_d$ by the hybrid control law $u(x,k,m)$ in the {\it obstacle-avoidance mode}. The first condition ({\it i.e.,} the velocity vector is tangent to the considered obstacle) is satisfied by construction for $m=\pm1$ since $\kappa(x,k,m)\in\mathcal{V}(c_k-x,\theta(x,k))$, as defined in Lemma \ref{lem6}. Let us show that the second condition is met by the hybrid control law $u(x,k,m)$ with respect to $x_d$ for $m\in\{-1,1\}$ and $k\in\mathbb{I}$ $\bigl(${\it i.e.,} $u(x,k,m)$ minimizes the deviation with respect to the nominal direction $(x_d-x)\bigr)$. Since the velocity vector $u(x,k,m)$ ensures a minimum angle with the nominal direction to the virtual destination $(x_k^m-x)$ given by $\angle(x_k^m-x,u(x,k,m))=\arccos(\frac{\kappa(x,k,m)^{\top}(c_k-x)}{\|\kappa(x,k,m)\|\|c_k-x\|})=\theta(x,k)\\-\beta(x,k,m)$, then one has to show that $\angle(x_d-x,u(x,k,m))\\=\theta(x,k)-\beta(x,k,0)$ $\bigl($or equivalently $\frac{\kappa(x,k,m)^{\top}(x_d-x)}{\|\kappa(x,k,m)\|\|x_d-x\|}=\cos(\theta(x,k)-\beta(x,k,0))\bigr)$ where  $\beta(x,k,0)=\angle(x_d-x,c_k-x)$. Hence,  $\frac{\kappa(x,k,m)^{\top}(x_d-x)}{\|\kappa(x,k,m)\|\|x_d-x\|}=\frac{\sin(\theta(x,k))}{\sin(\beta(x,k,m))}\bigl(\frac{(x_k^m-x)^{\top}(x_d-x)}{\|x_k^m-x\|\|x_d-x\|}-\frac{\sin(\theta(x,k)-\beta(x,k,m))}{\sin(\theta(x,k))}\frac{(c_k-x)^{\top}(x_d-x)}{\|c_k-x\|\|x_d-x\|}\bigr)$. Since, if $x(t_0^k,j_0^k)\notin\mathcal{L}(x_d,c_k)$, $x_k^m\in\mathcal{PL}(x_d,c_k,x(t_0^k,j_0^k))$  $\bigl($resp. if $x(t_0^k,j_0^k)\in\mathcal{L}(x_d,c_k)$, $x_k^m\in\mathcal{PL}(x_d,c_k,y)$, where $y\in\mathbb{R}^n\setminus\mathcal{L}(x_d,c_k)\bigr)$, and the {\it obstacle-avoidance} mode is selected such that the associated virtual destination is the closest to the position $x$, then $\angle(x_k^m-x,x_d-x)=\beta(x,k,m)-\beta(x,k,0)\geq 0$. Thus, $\frac{\kappa(x,k,m)^{\top}(x_d-x)}{\|\kappa(x,k,m)\|\|x_d-x\|}=\frac{\sin(\theta(x,k))}{\sin(\beta(x,k,m))}\bigl(\cos(\beta(x,k,m)-\beta(x,k,0))-\frac{\sin(\theta(x,k)-\beta(x,k,m))}{\sin(\theta(x,k))}\cos(\beta(x,k,0)\bigr)=\cos(\theta(x,k)-\beta(x,k,0))$. The velocity vector $u(x,k,m)$ is tangent obstacle $\mathcal{O}_k$ and ensures a minimum angle with the nominal direction $(x_d-x)$ for all $x\in\mathcal{F}_k^m\cap\mathcal{B}(c_k,r_k+\bar{r}_k-\epsilon)$, $m\in\{-1,1\}$ and $k\in\mathbb{I}$. Therefore, one can conclude that the control $u(x,k,m)$ generates local optimal obstacle avoidance maneuvers.
\bibliographystyle{ieeetr}
\bibliography{old/TRO_arxiv_version}
\thispagestyle{empty}
\end{document}